\documentclass[letterpaper]{article} % DO NOT CHANGE THIS
\usepackage{aaai24}  % DO NOT CHANGE THIS
\usepackage{times}  % DO NOT CHANGE THIS
\usepackage{helvet}  % DO NOT CHANGE THIS
\usepackage{courier}  % DO NOT CHANGE THIS
\usepackage[hyphens]{url}  % DO NOT CHANGE THIS
\usepackage{graphicx} % DO NOT CHANGE THIS
\usepackage{natbib}  % DO NOT CHANGE THIS AND DO NOT ADD ANY OPTIONS TO IT
\usepackage{caption} % DO NOT CHANGE THIS AND DO NOT ADD ANY OPTIONS TO IT
\usepackage{algorithm}
\usepackage[noend]{algorithmic}
\usepackage{subfigure}
\usepackage{graphicx}
\usepackage{amsthm}
\usepackage{amsmath}
\usepackage{amssymb}
\usepackage{booktabs}
\usepackage{multirow}
\usepackage{xcolor}

\usepackage{newfloat}
\usepackage{listings}
\DeclareCaptionStyle{ruled}{labelfont=normalfont,labelsep=colon,strut=off} % DO NOT CHANGE THIS
\floatstyle{ruled}
\newfloat{listing}{tb}{lst}{}
\floatname{listing}{Listing}
\newtheorem{thm}{\bf Theorem}
\newtheorem{lemma}{Lemma}

\title{Mitigating the Impact of False Negatives in Dense Retrieval \\ with Contrastive Confidence Regularization}
\author{
    Shiqi Wang,
    Yeqin Zhang,
    Cam-Tu Nguyen\thanks{Corresponding authors.}
    }
\affiliations{
    National Key Laboratory for Novel Software Technology, Nanjing University, China\\
    School of Artificial Intelligence, Nanjing University, China\\
    \{wangsky,zhangyeqin\}@smail.nju.edu.cn, \{ncamtu\}@nju.edu.cn\\
}

\begin{document}

\maketitle

\begin{abstract}
In open-domain Question Answering (QA), dense retrieval is crucial for finding relevant passages for answer generation. Typically, contrastive learning is used to train a retrieval model that maps passages and queries to the same semantic space. The objective is to make similar ones closer and dissimilar ones further apart. However, training such a system is challenging due to the false negative issue, where relevant passages may be missed during data annotation. Hard negative sampling, which is commonly used to improve contrastive learning, can introduce more noise in training. This is because hard negatives are those closer to a given query, and thus more likely to be false negatives. To address this issue, we propose a novel contrastive confidence regularizer for Noise Contrastive Estimation (NCE) loss, a commonly used loss for dense retrieval. Our analysis shows that the regularizer helps dense retrieval models be more robust against false negatives with a theoretical guarantee. Additionally, we propose a model-agnostic method to filter out noisy negative passages in the dataset, improving any downstream dense retrieval models. Through experiments on three datasets, we demonstrate that our method achieves better retrieval performance in comparison to existing state-of-the-art dense retrieval systems.

% This is particularly common in practice due to a large number of passage candidates and the complexity of the input queries.
% Recently, there has been a growing interest in augmenting LLM with a retrieval model as it can help reduce hallucinations, thus improving outputs generated by LLM. Dense retrieval methods are the most popular retrieval method due to their high performance. The dense retrieval model is typically trained  on labeled data, in which a query or a question is annotated with relevant evidence (texts). However, due to the large size of candidate passages, it is possible that not all pieces of evidence are labeled as positive during the data annotation, resulting in a large number of unlabeled positives in the corpus. On the other hand, introducing sampled hard negative examples during training has been shown its great help in learning a good retriever, which selects irrelevant samples close to given queries/ questions, and such a mechanism makes it more likely to select false negative samples, leading to ineffective training for the retrieval model. 

\end{abstract}
\section{Introduction}
Text retrieval involves searching for relevant information in vast text collections based on user queries. Efficient and effective methods for this task have revolutionized how we interact with information systems. Recently, there has been growing interest in augmenting large language models (LLMs) with text retrieval for question answering (QA)  \cite{rag, realm,  re2g, DBLP:conf/icml/BorgeaudMHCRM0L22, fu2022doc2bot, re3g}. These approaches harness retrieval models to obtain external knowledge and ground LLM outputs, reducing hallucinations and the need for frequent LLM updates.  Interestingly, augmenting an LM with a retrieval helps reduce the number of parameters required to achieve similar performance as larger LMs \cite{mialon2023augmented}.

%Interestingly, researchers have found that smaller retrieval-augmented LLMs often outperform their larger counterparts, balancing optimal QA performance with computational efficiency 

Text retrieval methods can be broadly categorized into two main approaches: sparse and dense retrievals. Sparse methods, such as BM25, exploit the frequency of words to measure the relevance between a passage and a query. While efficient, these methods often fall short of capturing intricate relationships and contextual nuances of language. In contrast, dense retrieval methods aim to learn meaningful representations from the semantic content of passages and queries effectively. These models can be trained based on a pretraining model (e.g. BERT, RoBERTa) as well as fine-tuned for downstream QA tasks \cite{DBLP:conf/nips/LewisPPPKGKLYR020}, offering easy integration. In addition, it is possible to apply approximate nearest neighbors (ANN) with dense retrieval \cite{ance} for efficient retrieval. %As a result, dense retrieval has become the preferred approach lately. 

% One key challenge in retrieval is negative sampling. On the one hand, challenging negative samples can serve as a catalyst for enhancing performance. Conversely, given the sheer volume of potential candidates, most datasets are prone to a high likelihood of omitting positive samples.  On the other hand, due to the vast number of candidates, most datasets have a high probability of positive sample omission, which can result in the collection of false negative samples. These false negative examples pose a significant bottleneck for dense retrieval performance, and a trade-off exists between selecting harder negative examples and the increased likelihood of collecting false negatives. The aim of \cite{rocketQA, simans, DBLP:conf/emnlp/Ni0D21} are from different perspectives, alleviating the problem of false negative examples and ultimately achieve improved results.

This paper focuses on dense retrieval, where contrastive learning is often employed to train passage and query encoders. The core principle of contrastive learning is to encode passages and queries such that relevant passages are closer to their corresponding query in the embedding space, while irrelevant passages are farther away. To train such encoders, we need a labeled dataset with queries annotated with relevant passages (positive samples). However, due to the vast number of candidate passages and the complexity of questions, it is common for annotators to miss relevant information (texts) during data preparation, leading to unlabeled positive examples (false negatives) in the training set. Recent studies support this assumption. For instance, \citet{mfnc} found that over half of 50 answerable questions from the IIRC dataset \cite{ferguson-etal-2020-iirc} had at least one missing piece of evidence. Similarly, \citet{rocketQA} manually reviewed top-retrieved passages not labeled as positives in MSMARCO \cite{dataset-ms} and detected a 70\% false negative rate. On the other hand, it is essential to sample hard negatives for effective contrastive learning. Here, hard negatives refer to passages obtained from the top results of a pre-trained dense retrieval model or BM25. Unfortunately, hard negative sampling is susceptible to higher false negative rates in noisy datasets because such negative samples are more likely to be mislabeled ones. Therefore, mitigating the impact of false negatives can potentially improve the performance of dense retrieval.

% Conversely, given the sheer volume of potential candidates, most datasets are prone to a high likelihood of omitting positive samples. This can inadvertently lead to the accumulation of false negative samples. These false negatives present a substantial impediment to the optimization of dense retrieval performance. A delicate balance must be struck between the selection of more difficult negative examples and the escalating risk of gathering false negatives. \cite{rocketQA, simans, mfnc}, each from their unique perspectives, strive to mitigate the problem of false negative examples. They aim to reduce the incidence of false negatives, thereby improving the overall effectiveness and accuracy of retrieval systems.

Several strategies have recently emerged to address the problem of false negatives. \citet{rocketQA} use a highly effective but inefficient reranker based on a cross-encoder to identify high-confidence negatives as true negatives, which were then used to train the retrieval model. \citet{mfnc} leverage answers in the downstream QA task and design several heuristics to detect valid contexts such as lexical overlapping (between a gold answer and a candidate passage). Recently, \citet{simans} suggests selecting samples that are highly similar to positive samples but not too close to the query. These samples are considered informative negatives and unlikely to be false negatives. Despite the recent progress, these current methods are primarily based on heuristics and lack a theoretical guarantee.

This paper formalizes the problem of training dense retrieval with false negatives into the peer loss framework \cite{peer-loss, cores2}, a theoretical sound approach to learning with label noise. We extend this framework by developing a confidence regularizer for NCE, a commonly used loss for training dense retrieval models. Our regularized loss function increases the model's confidence and proves to be robust against false negatives. By encouraging confident scoring, we prevent the model from overfitting to noise, resulting in a more robust retrieval model. We then propose a new passage sieve algorithm, that makes use of a confidence regularized retrieval model to select true hard negatives. The clean dataset after the passage sieve is then used to train a stronger retrieval model. We prove that our method can successfully filter out false negatives from hard negatives under mild assumptions. Through experiments on three datasets, we demonstrate that our method achieves better retrieval performance compared to existing state-of-the-art dense retrieval systems that rely on heuristic false negative filtering. Supplementary materials (the appendix, codes) can be found in our  GitHub\footnote{https://github.com/wangskyGit/passage-sieve}.

\section{Related Works}
\subsection{Dense Retrieval} The dual-encoder (two-towers or biencoder) architecture \cite{10.1145/2505515.2505665, sentence-bert} is the common choice for dense retrieval thanks to its high efficiency. However, vanilla dual encoders have several challenges such as the limited expressiveness compared to cross-encoders, and the suboptimal performance due to non-informative negative samples. As a result, various solutions have been introduced to improve vanilla dual encoders from different perspectives such as knowledge distillation \cite{rocketqav2, ernie}, lightweight interaction models \cite{khattab2020colbert, humeau2019poly}, sophisticated training procedure \cite{AR2,rocketQA,rocketqav2},  negative sampling strategies \cite{dpr,ance,rocketQA,simans}. In this paper, we focus mostly on negative sampling strategies, particularly targeting the false negative issue. Unlike the closely related works \cite{rocketQA, simans, mfnc}, which exploit heuristic strategies,  our method leverages the peer-loss approach \cite{peer-loss}, effectively combining practical application with a theoretically-informed perspective.

\paragraph{}

\subsection{Label-Noise Robust Machine Learning}
Developing machine learning models that are robust against label noise is important for supervised learning. Existing methods tackle label noise based on the type of noise, such as random noise \cite{natarajan2013learning,manwani2013noise}, class-dependent noise \cite{liu2015classification,patrini2017making,yao2020dual}, or instance-dependent noise \cite{second-order,cores2,xia2020part,yang2022estimating,hao2022model}. Unfortunately, these methods are mainly designed for multi-class classification, and thus cannot be directly applied to our task.

% It is desirable to develop machine learning models to be robust against label noise existed in training data, a common issue for supervised-learning. Based on the assumption of the noise type, current methods can be divided into solutions for random noise \cite{natarajan2013learning,manwani2013noise},  class-dependent noise \cite{liu2015classification,patrini2017making,yao2020dual}, or instance-dependent noise \cite{second-order,cores2,xia2020part,yang2022estimating,hao2022model}. However, most of them are designed for multi-class classification and can not be directly applied to our dense passage ranking task.

%\paragraph{Compared with Debiased Contrastive Loss Debiased contrastive learning}
Also relevant to our work are \cite{chuang2020debiased,robinson2020contrastive} which consider the issue of bias in contrastive learning for multi-class classification. %, which is different from noisy supervised contrastive learning for query-dependent ranking in our work. 
These studies rely on the assumption of a fixed uniform noise probability across different classes (i.e. queries in our case) to approximate the positive and negative distributions. In contrast, our work concerns noise in supervised contrastive learning for query-dependent ranking, where the distributions of positives and negatives are not the same for different queries. For example, it is more likely for queries with high recall to be associated with false negatives. In other words, the noise probability is not uniform across queries.

\section{Preliminaries}
\subsection{Problem Formalization} Let $C$ be a collection of textual passages, and $\tilde{\mathcal{D}} =\{(q_n,p_n^+,\tilde{\mathcal{N}}_{q_{n}})\}$ indicates the noisy set of tuples, each consists of a query $q_n$, an annotated positive $p_n^+\in C$, and a set of sampled negatives $\mathcal{N}_{q_{n}}\subset C$ with possible false negatives. Our objective is to mitigate the impacts of such false negatives and learn a robust retrieval model that can closely match the model trained on the clean dataset $\mathcal{D}=\{(q_n,p_n^+,\mathcal{N}_{q_{n}})\}$ without false negatives.

% Provided with a query $q$ and a corpus $C$, the objective of the retrieval task is to locate a collection of passages that are relevant to the query within C. These identified passages subsequently function as input for downstream models, for example, prompts used by large language models. Different from traditional methods that use the sparse index, dense retrieval methods train encoders from given dataset $D$ with the query $q$, positive passage $p^+$, and negative passage lists $\mathcal{N}_q$. And it then calculates the similarity of the encoded embeddings. Datasets with false negatives in the negative passage lists are termed as noisy datasets and represented as $\tilde{\mathcal{D}} ={(q_n,p_n^+,\tilde{\mathcal{N}}_{q_{n}})}$, while datasets without false negatives are termed as clean datasets and represented as $\mathcal{D}$. The object of the false negative robust dense retrieval model is to learn the retrieval model that retrieval the same passages given query among corpus $C$ with noisy dataset $\tilde{\mathcal{D}}$ as with clean dataset $\mathcal{D}$.

\subsection{Dual-Encoders for Dense Retrieval} A question or a passage is encoded as dense vectors separately, and the similarity is determined as the dot product of the encoder outputs.
\begin{equation}
    sim(q, p) = <E_{qry}(q), E_{psg}(p)>
\end{equation}
where $E_{qry}$, $E_{psg}$ represent distinct encoders that map the query and the passage into dense vectors, and $<.>$ is the similarity function such as dot product, cosine or Euclidean distance. The encoders are often built based on a pre-trained language model such as BERT-based \cite{dpr, me-bert, ance, DPR-PAQ, mvr}, ERNIE-based \cite{rocketQA, rocketqav2, AR2} or RoBERTa-based \cite{DPR-PAQ} etc. Since passages and queries are encoded separately, passage embeddings can be precomputed and indexed using Faiss \cite{faiss} for efficient search. 

Contrastive learning is commonly used to train dual encoders \cite{dpr, ance, rocketQA, simans}. Typically,  it is assumed that we have access to the clean training set $\mathcal{D}=\{(q_n,p_n^+,\mathcal{N}_{q_{n}})\}$. Using this training set, we measure the NCE (Noise-Contrastive Estimation) loss for a given  query $q_n$ as follows:
\begin{equation}
\begin{aligned}
    \ell_{NCE}(p_n^+,q_n)&=-\ln(f(p_n^+,q_n))\\
    &=-\ln(\frac{e^{sim(p_n^+,q_n)}}{\sum_{p_i^- \in \mathcal{N}_{q_n}}e^{sim(p_i^-,q_n)}+e^{sim(p_n^+,q_n)}})
    \label{eq:nce}
\end{aligned}  
\end{equation}
\noindent The total loss function can be calculated as follows:
\begin{equation}
  \frac{1}{N}\sum_{n\in [N]}\ell_{NCE}(p_n^+,q_n)  
  \label{eq:nce_total}
\end{equation}
where $N$ is the total number of queries in the dataset, and $[N]=\{1,2,\ldots, N\}$. 

Negative sampling aims to select samples for $\mathcal{N}_{q_n}$ and plays an important role in learning effective representations with contrastive learning. In the context of dense retrieval, two common strategies for negative sampling are in-batch negatives and hard negatives \cite{dpr,ance,rocketQA}. In-batch negatives involve selecting positive passages from other queries in the same batch as negative samples. Generally, increasing the number of in-batch negatives improves dense retrieval performance. On the other hand, hard negatives are usually informative samples that receive a high similarity score from another retrieval model (e.g., BM25, a pre-trained DPR) \cite{dpr}. Hard negatives can result in more effective training for DPR, yet including such samples may exaggerate the issue of false negatives. %It is important to note that there are possibly multiple positive passages.  We, however, follow previous studies \cite{rocketQA} and assume that only one is annotated for one query in the noisy training set.

\subsection{Peer Loss and Confidence Regularization}
The problem of learning with label noise has been extensively researched in the context of classification tasks \cite{peer-loss,xia2020part, second-order,cores2,yang2022estimating,hao2022model}. Recently, two effective methods called Peer Loss \cite{peer-loss} and its inspired Confidence Regularizer \cite{cores2, second-order} have been introduced to train robust machine learning models. One advantage of these methods is that they can work without requiring knowledge of the noise transition matrix or the probability of labels being flipped between classes.

The concept of peer loss is initially introduced for the binary classification problem. In this setting, we use $x$ to indicate an instance (a feature vector), $y \in \{-1,1\}$ and $\tilde{y}\in \{-1,1\}$ to represent the clean label and noisy labels, respectively. For each sample $(x_n,\tilde{y}_n)$, the peer loss is then defined for cross-entropy loss as follows:
\begin{equation}
    \ell_{PL}(g(x_n),\tilde{y}_n) = \ell_{CE}(g(x_n),\tilde{y}_n)-\ell_{CE}(g(x_{n1}),\tilde{y}_{n2})
    \label{eqn:original_pl}
\end{equation}

\noindent where $g$ indicates the classification function, $x_{n1}$ and $\tilde{y}_{n2}$ are derived from different, randomly selected peer sample for $n$. Here, the first term measures the loss of the classifier prediction, whereas the second term penalizes the model when it excessively agrees with the incorrect or noisy labels \cite{peer-loss}.

Inspired by peer loss, \citet{cores2} develops $\text{CORES}^2$, which extends the framework for multi-class classification and uses the first-order statistic instead of randomly selecting peer samples. Specifically, the new loss in $\text{CORES}^2$ is defined as follows:
\begin{equation}
    \begin{aligned}
         &\ell(g(x_n),\tilde{y}_n) =
         \ell_{CE}(g(x_n),\tilde{y}_n)-\beta \mathbb{E}_{\tilde{D}_{\tilde{Y}}}[\ell_{CE}(g(x_n),\tilde{Y})]
    \end{aligned}
    \label{eqn:cores}
\end{equation}
where $\beta$ is a hyper-parameter, $\tilde{Y}$ denotes the random variable corresponding to the noisy label and $\tilde{D}_{\tilde{Y}}$ indicates the marginal distribution of $\tilde{Y}$. It has been shown that learning with an appropriate $\beta$ will make the loss function robust to instance-dependent label noise with theoretical guarantee \cite{cores2}. % Although these methods have shown great power in dealing with noisy data in both practice and theory, they are designed for supervised learning with cross-entropy loss, and can not be used for problems using contrastive loss like NCE loss function in the dense retrieval problem.

\section{Contrastive Confidence Regularizer\label{sec:CCR}}

We can adopt the above confidence regularization by introducing a binary label $y$ associated with each pair of queries and passages, where $y=+1$ indicates the positive pair and $y=-1$ vice versa. Subsequently, $\text{CORES}^2$ can be used with pairwise cross-entropy to mitigate the impacts of false negatives on training the retrieval model. Unfortunately, the cross-entropy loss is not as effective as the NCE loss (Eq. \ref{eq:nce_total}) for dense retrieval since the latter can learn a good ranking function by contrasting a positive sample over a list of negatives \cite{dpr}. As a result, we aim to extend the framework of peer loss and tailor the confidence regularization to the NCE loss. 

% The task of training dense retrieval models with false negatives can be seen as a special case of learning with label noise. Our problem, however, exhibits several distinct characteristics that need our attention. Firstly, the original confidence regularizer in $\text{CORES}^2$ calculates the estimation of the cross-entropy loss of different labels. In contrast, our new contrastive confidence regularizer calculates the estimation of NCE loss among different passages given the query in the noisy dataset $\tilde{\mathcal{D}}$. For a more specific comparison,  let's consider y=1 as the label for positive pairs and y=-1 as the label for negative pairs in dense passage retrieval. If we wish to calculate the original confidence regularize in \cite{cores2}, we must have $\ell(p_n,q_n;y=1)$ and $\ell(p_n,q_n;y=-1)$ at the same time, however, the later one is missing in dense passage retrieval or any other contrastive learning methods. \torevise{Second, given the NCE loss \ref{eq:nce} is defined for ``detecting a positive sample from a list of positive and negative samples'', and so we have loss associated with positive samples only.} Therefore, the original confidence regularizer cannot be applied, let alone its theoretical basis is no longer valid with a totally different loss function. 

To adopt the peer-loss framework, we extend NCE loss to measure loss values associated with negative pairs $(p_i^-,q_n)$ besides positive ones. It is noteworthy that the original NCE loss (Eq. \ref{eq:nce_total}) incorporates all positive and negative pairs as normalization factors but only calculates loss values for positive pairs \((p^+_n, q_n)\) while disregarding the negative pairs. Formally, we use $p$ without superscript as a generic term for positive passage $p_n^+$ and negative passage $p_i^- \in \mathcal{N}_{q_n}$, and define a general NCE loss as follows:
\begin{equation}
\begin{aligned}
    \ell_{NCE}(p,q_n)&=-\ln(f(p,q_n))\\
    &=-\ln(\frac{e^{sim(p,q_n)}}{\sum_{p_i^- \in \mathcal{N}_{q_n}}e^{sim(p_i^-,q_n)}+e^{sim(p_n^+,q_n)}})
\end{aligned}
\end{equation}
% where p could be either positive or negative passages for query $q_n$. By this convention, the peer-loss framework can be readily applied by introducing randomly selected pairs $(p_{n_1}, q_{n_1})$ (${n_1}\neq {n}$) together with $(p_n,q_n)$ as the peer sample to regularize the NCE loss. We then obtain the contrastive peer loss as follows:
where p could be either positive or negative passages for query $q_n$. By this convention, the peer-loss framework can be applied by introducing randomly selected pairs $(p_{n_1}, q_{n_1})$, $(p_{n_2}, q_{n_2})$ (${n_1}\neq {n_2}$) as peer samples to regularize the NCE loss. We then obtain the contrastive peer loss as follows:
\begin{equation}
\ell_{PL}(p_n^+,q_n) = \ell_{NCE}(p_n^+,q_n)-\ell_{NCE}(p_{n_1},q_{n_2})
\label{eqn:contrastive_pl}
\end{equation}

\noindent While our regularized loss appears similar to the original peer-loss outlined in Eq. \ref{eqn:original_pl}, the major difference lies in the structure of the loss functions. Specifically, the first term of Eq. \ref{eqn:contrastive_pl} is calculated only for positive pairs whereas the first term of Eq. \ref{eqn:original_pl} involves both positive ($\tilde{y}=+1$) and negative ($\tilde{y}=-1$) samples.

Similar to $\text{CORES}^2$, we introduce $P_{n_1}$ and $Q_{n_2}$ as the random variables corresponding to the peer passage $p_{n_1}$ and the peer query $q_{n_2}$, respectively. Let $\tilde{\mathcal{D}}_P$ be the distribution of $P_{n_1}$ given distribution $\tilde{\mathcal{D}}$. Note that $Q_{n_2}$ is a uniform random variable, or $\mathbb{P}(Q_{n_2}=q_{n'}|\mathcal{\tilde{D}})=1/N$, the contrastive peer loss has the following form in expectation:
\begin{equation*}
    \begin{aligned}
        &\frac{1}{N}\sum_{n\in[N]}\mathbb{E}_{\tilde{\mathcal{D}}_{P_{n_1},Q_{n_2}}}\left[\ell_{NCE}(p_n^+,q_n)-\ell_{NCE}(P_{n_1},Q_{n_2})\right]\\
        =&\frac{1}{N}\sum_{n\in[N]}[\ell_{NCE}(p_n^+,q_n)-\\
        &\sum_{n' \in[N]} \mathbb{P}   (Q_{n_2}=q_{n'}|\tilde{\mathcal{D}})\mathbb{E}_{\tilde{\mathcal{D}}_{P}}\left[\ell_{NCE}(P_{n_1},q_{n'})\right]\\
        = &\frac{1}{N}\sum_{n\in[N]}\left[\ell_{NCE}(p_n^+,q_{n})-\mathbb{E}_{\tilde{\mathcal{D}}_{P}}\left[\ell_{NCE}(P,q_{n})\right]\right]\\
        \approx &\frac{1}{N}\sum_{n\in[N]}\left[\ell_{NCE}(p_n^+,q_{n})-\mathbb{E}_{\tilde{\mathcal{D}}_{P|q_{n}}}\left[\ell_{NCE}(P,q_{n})\right]\right]
    \end{aligned}
\end{equation*}
The last approximation equation is obtained because when considering the batch training in dense retrieval, $P_{n_1}$ is drawn from in-batch passages. In addition, all the passages in batch form the conditional passage distribution of query $q_n$.  From this derivation, we can get the new noise robust contrastive loss function (i.e. $\ell_{RCL}$) with contrastive confidence regularizer denoted by $\ell_{CCR}$ as follows:
\begin{equation}
\begin{aligned}
    \ell_{CCR}(p_n^+,q_n) &= \mathbb{E}_{\tilde{\mathcal{D}}_{P|q_{n}}}[\ell_{NCE}(p,q_{n})]\\
    \ell_{RCL}(p_n^+,q_n) &=\ell_{NCE}(p_n^+,q_n)-\beta*\ell_{CCR}(p_n^+,q_n) \label{l_RCL}
\end{aligned}
\end{equation}

\noindent In the following, we first empirically prove that our regularized loss function makes the retrieval model more confident, hence being more robust against false negatives. We then present the main theories that guarantee the robustness of our regularized NCE loss.

\subsection{Analysis on Simulated Data} 
\begin{figure}[t]
    \centering
    \subfigure[$\beta=0$]{\includegraphics[width=0.48\columnwidth]{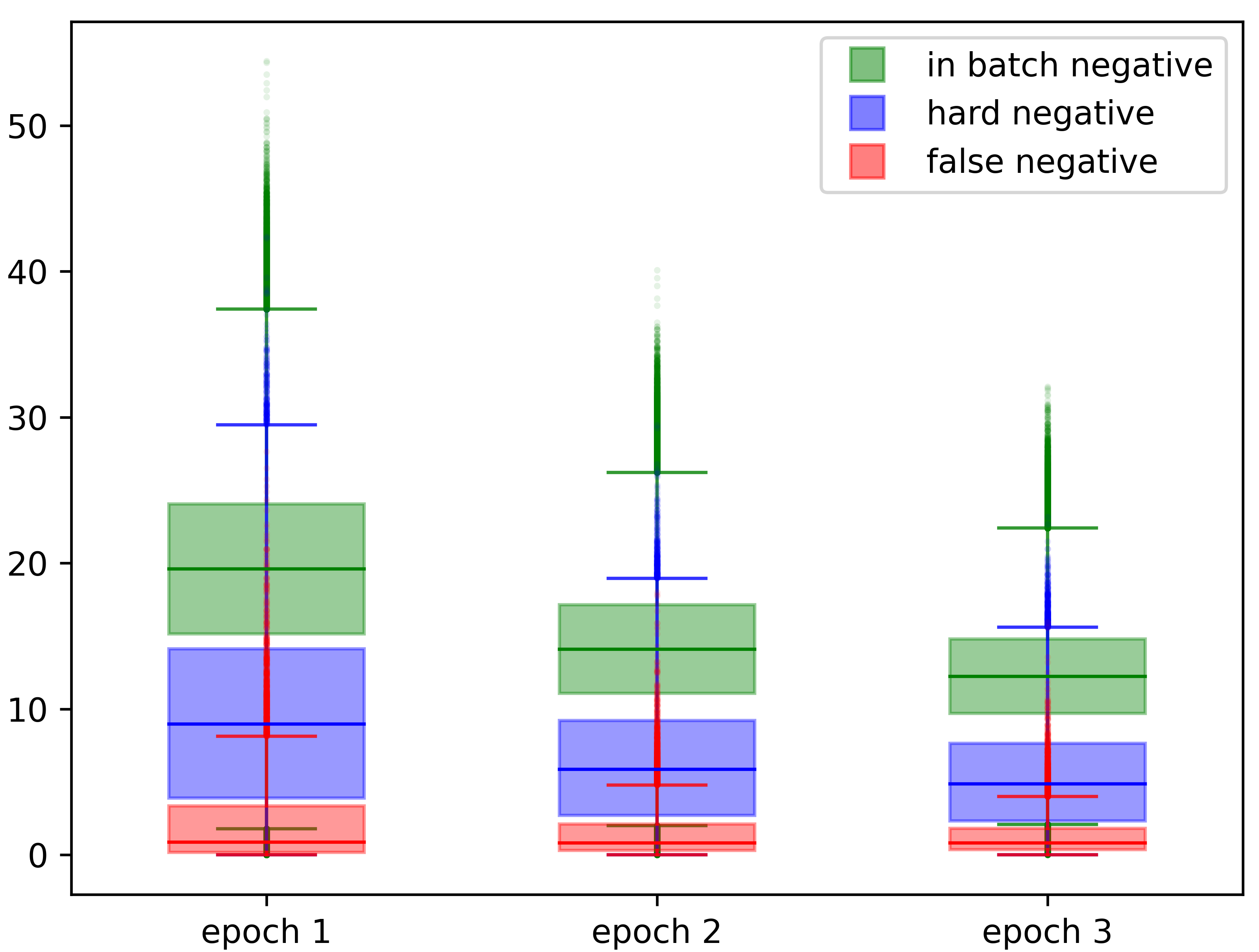}}
    \subfigure[$\beta=0.05$]{\includegraphics[width=0.48\columnwidth]{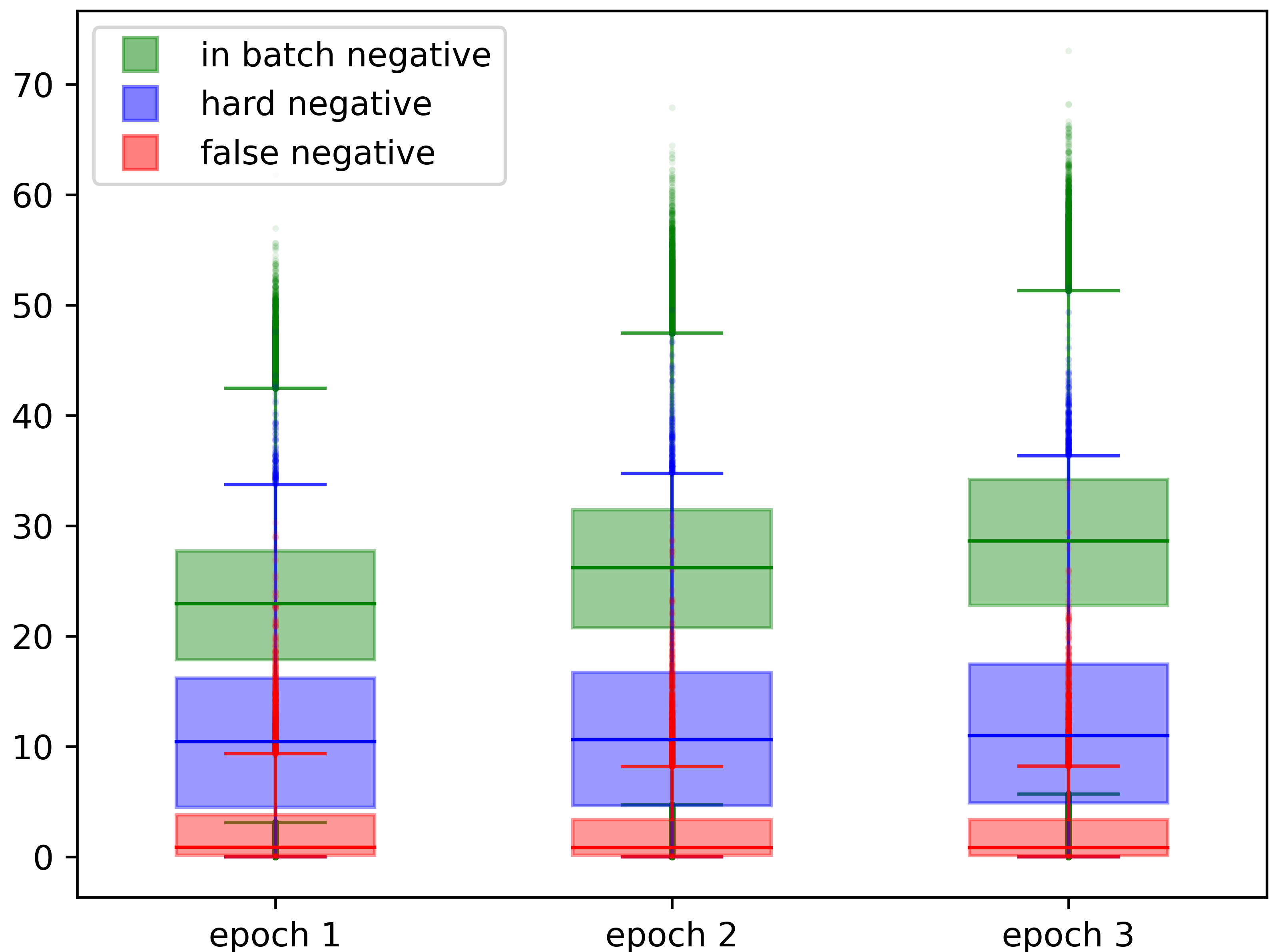}}

    \subfigure[$\beta=0.1$]{\includegraphics[width=0.48\columnwidth]{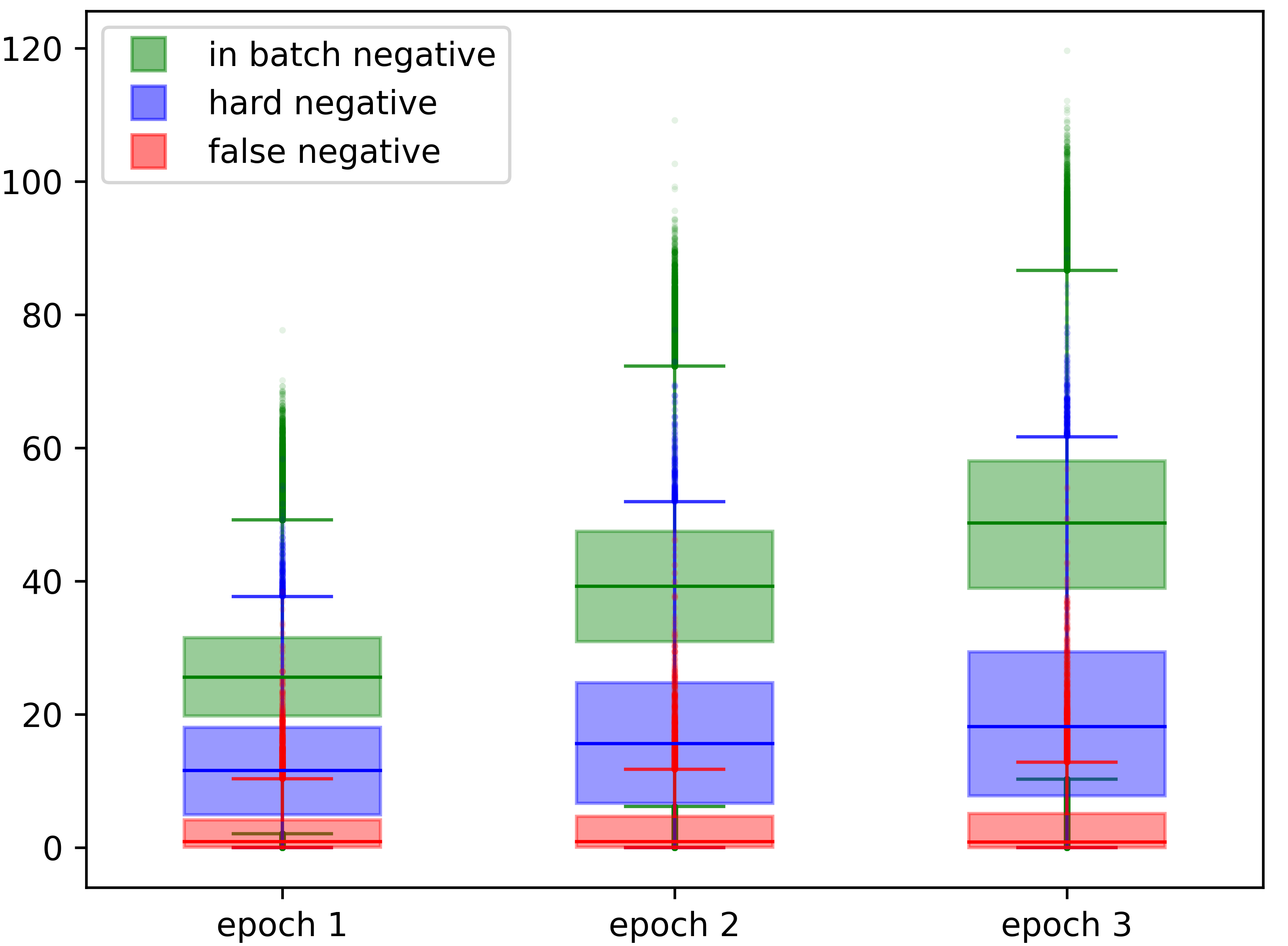}}
    \subfigure[$\beta=0.5$]{\includegraphics[width=0.48\columnwidth]{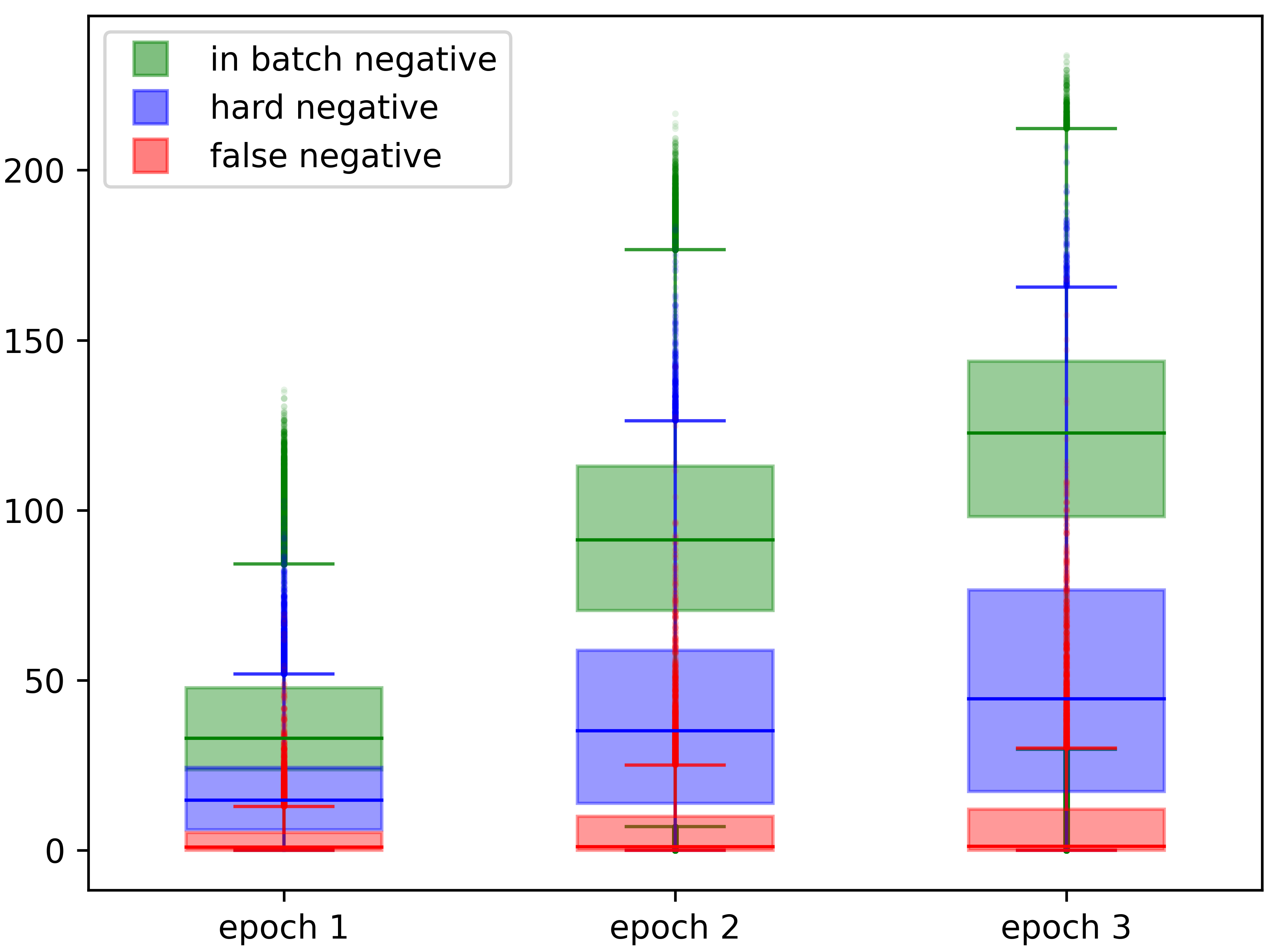}}
    \caption{The effect of the contrastive confidence regularizer. Four figures show analysis from four experiments with all the same settings but with different values of beta. All of them are continuously trained based on the same pre-trained vanilla DPR. The y-axis value represents the NCE loss value between queries and different types of passages. It shows that normal DPR training will cause different distributions to be squeezed together while $\ell_{CCR}$ has the potential to help distinguish false negatives from real negatives.}
    \label{fig:simulation}
\end{figure}

 The RCL loss (eq. \ref{l_RCL}) is minimized by making the first term smaller and the expectation term bigger. In other words, we pull the loss associated with positive passages further from the average (the expectation) loss, subsequently making the model more confident in predicting positive passages. Intuitively, as label noise distorts the learning signal in clean data, the model trained on noisy datasets often fits noises, hence becoming less confident \cite{cores2}. By making the model more confident, we can partially reverse it and obtain a more robust retrieval model. We verify this intuition by conducting a simulation on the Natural Question (NQ) dataset \cite{dataset-NQ}. Specifically, we randomly convert some positive passages to false negatives and observe the effects of contrastive confidence regularizer $\ell_{CCR}$ on the distributions of the similarity scores in different groups of passages including false negatives, hard negatives and in-batch negatives. The experimental results in Figure \ref{fig:simulation} indicate that if we do not include the $\ell_{CCR}$ term (i.e., $\beta=0$), the distributions are close to each other. On the other hand, incorporating the $\ell_{CCR}$ term makes it easier to separate these distributions.  It should be noted that a small overlapping between the false negative and hard negative distributions is expected. This is because while all samples in the simulated ``false negatives'' category are certainly positive passages, there exist (unknown) false negatives in the ``hard negatives'' category.

% This is consistent in $\text{CORES}^2$. By making the model $f$ more confident, it would give scores (the value after softmax) for a pair $(p,q)$ either close to 1 or 0 and avoid vague judgment. And so false negatives will be pushed away from the true negative cluster and help the loss function get the same classifier as with clean datasets. 

\subsection{Theoretical Analysis}
Theoretical analysis shows that our RCL loss enjoys similar properties with $CORES^2$. The detailed proof can be found in the Appendix, where the main idea is that we introduce pseudo-labels to bridge NCE loss and cross-entropy loss and follow a similar proof sketch with \cite{cores2}. The main result is summarized in the following theorem.

% The proposed contrastive confidence regularizer is significantly different from the original one, and thus new theory analysis is required. The detailed proof can be seen in the appendix. In short, the proof follows the vein of the original paper but with different content. And the key to the proof is to introduce the idea of pseudo-label into the NCE loss and bridge the contrastive loss and cross-entropy loss together. Finally, we will get the following theorem which gives us a theory guarantee for the proposed robust contrastive loss.

\begin{thm}
    With the assumption that the possibility of a noisy pair is smaller than a clean pair and a suitable selection of $\beta$, we have: minimizing $\mathbb{E}_{\tilde{\mathcal{D}}}[\ell_{NCE}(p^+,q)-\beta*\ell_{CCR}(p^+,q)]$ is equivalent to minimizing $\mathbb{E}_\mathcal{D}[\ell_{NCE}(p^+,q)]$
\end{thm} 

Theorem 1 indicates that the contrastive confidence regularizer effectively addresses the adverse effect of false negatives on the loss function. Specifically, we can decompose the regularized loss so that the impact of false negatives is mostly captured by the product of $\ell_{CCR}$ and a scaling factor (the last term in Eq.\ref{eq:e_rcl}, Appendix). Here, the scaling factor is intuitively related to the difference between the relative clean rate of a sample (compared to the average clean rate) and $\beta$. When $\beta$ is large enough, this scaling factor is negative, thus reversing the (term) gradient associated with instances with labeled noise. Note that $\beta$ should not be too large, otherwise it will affect learning
from clean data. This raises a crucial question regarding the existence of parameter $\beta$. Analyzing the existence of $\beta$ becomes challenging when the nature of the noise is uncertain. Fortunately, in the context of false negatives in dense passage retrieval, we can derive the following theorem.

% REPEATED THE THEORY: With a suitable $\beta$, minimizing the first-order statistics of the robust contrastive loss in the noisy dataset $\tilde{\mathcal{D}}$ is equivalent to that of the clean dataset $\mathcal{D}$. And thus adding the confidence regularizer term can help contrastive learning robust to any possible noise. And the parameter $\beta$ can control the effect of the regularizer. Pay attention that theorem 1 does not hypothesize what noise is in the noisy datasets. Both the false negatives problem in dense passage retrieval and the false positive problem in self-supervised image classification apply to the theorem. This demonstrates the universality of the proposed robust contrastive loss and can be easily applied to other applications using contrast loss

\begin{thm}
    When the assumption in Theorem 1 holds, the similarity function $sim$ is bounded (e.g. cosine similarity)  and there is no false positive in the dataset $\tilde{\mathcal{D}}$, the $\beta$ that satisfies Theorem 1 must exist and in the interval [0,1]. 
\end{thm}
Theorem 2 shows that under the setting of dense passage retrieval with false negatives, one can select a suitable $\beta$ between the interval [0,1]. However, this theorem only applies when the similarity function is bounded and there is no other source of noise besides false negatives. In practical scenarios, many dense passage retrieval algorithms employ dot-product similarity instead of cosine similarity. When the loss function is not bounded, and there is an excessive number of negative passages, the confidence regularizer can cause the model to overly optimize the loss associated with negative cases. Our experiments indicate that in such situations, the contrastive confidence regularizer still helps as long as a sufficiently small value for $\beta$ is selected. In general, as the batch size increases and in-batch negative sampling is utilized, we should decrease the value of $\beta$.

\section{Passage Sieve Method}\label{sec:passage_sieve}
\begin{algorithm}[t]
\caption{Passage Sieve Algorithm}
\label{algo:passage_sieve}
\textbf{Input}:  Noisy dataset $\tilde{\mathcal{D}}$ with false negatives\\
\textbf{Parameter}: Hyper-parameter $\beta$, learning rate $\alpha$, training epochs $T$\\
\textbf{Output}: Sieved dataset $\mathcal{D}^*$ %with confident false negatives
\begin{algorithmic}[1]
\STATE Get the pre-trained DPR model $\mathcal{M}$
\STATE Train $\mathcal{M}$ with robust contrastive loss for T epoch:
\FOR{$t \in [T]$}
  \FOR{$(\mathbf{q}, \mathbf{p^+},\mathbf{\mathcal{N}_{q}})$ \textbf{in} $\tilde{\mathcal{D}}$}
    %\STATE\COMMENT{$q$:query, $p^+$:positive passage of the query, $\mathcal{N}_{q}$:the set of hard negatives  of the query}
     %\STATE Define $D_{p|q} \triangleq \mathcal{N}_{q}\cup \{p^+\}$
     \STATE Calculate $\ell_{RCL}$ according to equation \ref{l_RCL} 
     \STATE Compute gradients $\nabla\leftarrow\nabla_{\mathcal{M}} \ell_{RCL}$
     \STATE Update the model parameters: $\mathcal{M} \leftarrow \mathcal{M} - \alpha \cdot \nabla$
  \ENDFOR
    \ENDFOR
\STATE Evaluate all the hard negatives:
\STATE Initialize empty dataset $\mathcal{D}^*$ 
\FOR{$(\mathbf{q}, \mathbf{p^+},\mathbf{\mathcal{N}_{q}})$ \textbf{in} $\tilde{\mathcal{D}}$}
    \STATE $\mathcal{N}_{q}^*\leftarrow []$ \COMMENT{list of selected hard negatives}
    \STATE $th = \frac{1}{|D_{p|q}|} \sum_{p\in D_{p|q}} \ell_{NCE}(p,q;\mathcal{M})$\COMMENT{Threshold}
    \FOR{$p^-$ \textbf{in} $\mathcal{N}_q$}
        \STATE $\ell_p^- \leftarrow \ell_{NCE}(p^-,q;\mathcal{M})$\;
        \IF{$\ell_p^- \geq th$}{
        \STATE Append $p^-$ to $\mathcal{N}_{q}^*$
        }
        \ENDIF
    \ENDFOR
    \STATE Append $(\mathbf{q}, \mathbf{p^+},\mathbf{\mathcal{N}_q^*})$ to $\mathcal{D}^*$ 
\ENDFOR
\STATE Return Sieved dataset $\mathcal{D}^*$ with confident false negatives
\end{algorithmic}
\end{algorithm}
The previously proposed regularization is helpful towards dense retrieval models based on contrastive NCE loss, but not applicable to sophisticated methods such as AR2 \cite{AR2}, which do not make use of NCE loss. As a result, we design a novel passage sieve algorithm, which can be used as pre-processing for the noisy datasets $\tilde{\mathcal{D}}$ to obtain relatively clean datasets for training.

Our method is presented in Algorithm \ref{algo:passage_sieve}. We first train DPR \cite{dpr}, a dual-encoder based retrieval model on the noisy dataset. We then refine the retrieval model using the contrastive confidence regularizer. Afterward, we identify the hard negative passages with a loss function value $\ell_{NCE}(p,q_n)$ higher than the average loss value $\frac{1}{|N_{q_n}|+1}\sum_{p \in N_{q_n} \cup \{p_n^+\} }\ell_{NCE}(p,q_n)$, and set them as confident negatives. Finally, we discard all other passages except for the confident negatives to obtain a ``clean'' dataset. It is noteworthy that we employ cosine similarity to satisfy the assumptions outlined in Theorem 2.

% Following a similar design as in the sample sieve, the false negative sieve problem is defined as follows:
% \begin{equation}
%     \begin{aligned}
%     \min_{f\in\mathcal{F},v\in\{0,1\}}\sum_{p_n,q_n}v_n[\ell(p_n,q_n;\tilde{y_n})+\ell_{CR}(p_n,q_n)-\alpha_n]\\
%     s.t.\quad \ell_{CR}(p_n,q_n)=-\beta\mathbb{E}_{\mathcal{D}_{p|q_n}}[\ell(p,q_n;-1)],\\
%     \alpha_n = \frac{1}{|\mathcal{N}_k|+1}\sum_{p_i \in \mathcal{N}_k\cup\{P_k\}}\bar{\ell}(p_i,q_n;-1)+\bar{\ell}_{CR}(p_n,q_n).
% \end{aligned}
% \end{equation}
% Similar to the dynamic sample sieve, problem (1) can be solved by applying alternate search iteratively. The major difference is the design of $\ell_{CR}$ and $\alpha_n$ compared to the original sample sieve method $\text{CORES}^2$ \cite{cheng2020learning}.  However, we will prove that our new regularizer can handle the retrieval problem and contrastive loss well with a theoretical guarantee. The key idea of the proof is to examine whether all the important theorems remain similar as in  $\text{CORES}^2$. 

\begin{lemma}
Algorithm \ref{algo:passage_sieve} ensures that a negative sample $p^-$ in hard negatives will NOT be selected into the sieved dataset $\mathcal{D}^*$ if its score $f(p^-,q)$ given by the model f (eq. \ref{eq:nce}) is more than a random guess, i.e. its similarity score after softmax is bigger than the average value $1/(|\mathcal{N}_q|+1)$.
\end{lemma}

% \begin{lemma} 
% For each query $q$, with the assumption that the set of true negatives $\{p|q,y=-1,\tilde{y}=-1\}$ is not empty, at least one true negative will be retained if all false negative passages $\{p|q,y=1,\tilde{y}=-1\}$ are sieved out.
% \end{lemma}

% Lemma 1 can be easily proved by the property that at least one element is bigger than the average (estimation) of the set (distribution). So, if all samples in the false negatives set are sieved out then at least one element in the true negatives set will be retained. Since the training process of retrieval often requires only one strong negative sample, this lemma allows us to consider only how to screen out all false negatives in the sample sieve process. Combining with Theorem 1 and Lemma 1, in order to get one true hard negative after sample sieve for the following passage retrieval training, we only need to prove that during the optimization of the regularized loss, all the positive passages (including false negative passages) tend to get $f(p,q)$ higher than the average $1/(|\mathcal{N}_k+1|)$.

Lemma 1 provides the sufficient condition for accurately selecting true negatives in Algorithm \ref{algo:passage_sieve}. In this algorithm, the DPR model is trained with robust contrastive loss and cosine similarity. According to Theorems 1 and 2, by selecting a suitable value of $\beta \in [0,1]$, minimizing $\mathbb{E}_{\tilde{D}}[\ell(p,q)+\ell_{CCR}(p,q)]$ is equivalent to minimizing $\mathbb{E}_\mathcal{D}[\ell(p,q)]$. This implies that false negatives are gradually given scores closer to the positive passage, and higher compared to hard negatives and in-batch negatives. Considering the abundance of true negatives, the scores assigned to false negatives will exceed random guesses at some point. Consequently, false negatives will be excluded from the sieved dataset in Algorithm \ref{algo:passage_sieve}, resulting in a clean dataset $\tilde{D}$.
% During the minimization of $\mathbb{E}_\mathcal{D}[\ell(p,q)]$ in the clean distribution, false negatives in $\tilde{\mathcal{D}}$ are labeled as positives. And the term $e^{sim(p_n,q_n)}$ of false negatives will appear only in the denominator of the contrastive loss function. Consequently, with infinite model capacity and enough examples, the score $f(p_n,q_n)$ of false negatives will be bigger than $1/(\mathcal{N}_k+1)$ and will NOT be selected as stated in lemma 1.  And in algorithm \ref{algo:passage_sieve}, we leverage the robust contrastive loss to train the DPR model and thus With Theorem 1 and Theorem 2, by minimizing $\mathbb{E}_{\tilde{D}}[\ell(p,q)+\ell_{CCR}(p,q)]$, we obtain the same effect as minimizing $\mathbb{E}_\mathcal{D}[\ell(p,q)]$, resulting in all true hard negatives being selected out. In conclusion, together with Lemma 1 Theorem 1 and Theorem2, with infinite model capacity and enough examples, the passage sieve process in algorithm \ref{algo:passage_sieve} shall select all the true hard negatives, and the noisy dataset $\tilde{D}$ shall be clean after the passage sieve.

Compared to vanilla DPR, DPR with CCR introduces additional computation associated with the expectation calculation, the second term in Eq. \ref{l_RCL}.  Fortunately, this can be approximated by taking the mean value of the NCE losses of all passages given the query. In addition, the calculation of NCE losses is dominated by the calculation of the similarity matrix (Eq. \ref{eq:nce}), which is also needed in vanilla DPR. Consequently, the contrastive confidence regularizer only slightly increases the time complexity when implemented appropriately. It is worth noting that when using retrieval models like DPR, ANCE, or RocketQA that employ NCE loss, using the robust contrastive loss in section \ref{sec:CCR} should be sufficient.

\section{Experiment}

\begin{table*}[t]
\centering
\resizebox{0.95\linewidth}{!}{%
\begin{tabular}{l|ccc|ccc|ccc}
\toprule
\multicolumn{1}{c|}{\multirow{2}{*}{\textbf{Method}}} & \multicolumn{3}{c|}{\textbf{NQ}}                    & \multicolumn{3}{c|}{\textbf{TQ}}                    & \multicolumn{3}{c}{\textbf{MS-pas}}                 \\
\multicolumn{1}{c|}{}                                 & \textbf{R@5} & \textbf{R@20} & \textbf{R@100} & \textbf{R@5} & \textbf{R@20} & \textbf{R@100} & \textbf{MRR@10} & \textbf{R@50} & \textbf{R@1k} \\ \midrule
DPR \cite{dpr}      & -              & 78.4            & 85.3             & -              & 79.3            & 84.9             & -               & -               & -               \\
ANCE \cite{ance}     & 71.8           & 81.9            & 87.5             & -              & 80.3            & 85.3             & 33.0            & 81.1            & 95.9            \\
COIL \cite{coil}      & -              & -               & -                & -              & -               & -                & 35.5            & -               & 96.3            \\
ME-BERT \cite{me-bert}   & -              & -               & -                & -              & -               & -                & 33.8            & -               & -               \\
Individual top-k \cite{individual-topk}  & 75.0           & 84.0            & 89.2             & 76.8           & 83.1            & 87.0             & -               & -               & -               \\
RocketQA \cite{rocketQA}   & 74.0           & 82.7            & 88.5             & -              & -               & -                & 37.0            & 85.5            & 97.9            \\
RDR \cite{RDR}       & -              & 82.8            & 88.2             & -              & 82.5            & 87.3             & -               & -               & -               \\
RocketQAv2 \cite{rocketqav2}    & 75.1           & 83.7            & 89.0             & -              & -               & -                & 38.8            & 86.2            & 98.1            \\
PAIR \cite{Pair}      & 74.9           & 83.5            & 89.1             & -              & -               & -                & 37.9            & 86.4            & 98.2            \\
DPR-PAQ \cite{DPR-PAQ}    & 74.2           & 84.0            & 89.2             & -              & -               & -                & 31.1            & -               & -               \\
Condenser \cite{condenser}     & -              & 83.2            & 88.4             & -              & 81.9            & 86.2             & 36.6            & -               & 97.4            \\
coCondenser \cite{cocondenser}     & 75.8           & 84.3            & 89.0             & 76.8           & 83.2            & 87.3             & 38.2            & -               & 98.4            \\
ERNIE-Search \cite{ernie}     & 77.0           & 85.3            & 89.7             & -              & -               & -                & \textbf{40.1}            & 87.7            & 98.2            \\
MVR \cite{mvr}     & 76.2           & 84.8            & 89.3             & 77.1              & 83.4               & 87.4                & -            & -            & -            \\
PROD \cite{lin2023prod}     & 75.6           & 84.7            & 89.6             & -              & -               & -                & 39.3            & 87.0            & 98.4            \\
COT-MAE \cite{cot-mae}     & 75.5           & 84.3            & 89.3             & -              & -               & -                & 39.4            & 87.0            & \textbf{98.7}            \\\midrule
AR2 \cite{AR2}         & 77.9           & 86.0            & 90.1             & 78.2           & 84.4            & 87.9             & 39.5            & 87.8            & 98.6            \\
AR2+passage sieve    & \textbf{79.2}              & \textbf{86.4}            & \textbf{90.7}              &\textbf{78.7}            & \textbf{84.7}            & \textbf{88.2}            &  \underline{39.8}               &    \textbf{88.3}         &      \underline{98.6}          \\
\bottomrule
\end{tabular}%
}
\caption{Performance of passage sieve with the robust contrastive loss on NQ, TQ test sets, and MS-pas development set. The results of the baselines are from the original papers. All results with AR2 as the backbone are obtained from training AR2 with the same batch size and the same number of hard negatives as in \cite{AR2}.}
%\caption{Performance of passage sieve with the robust contrastive loss on NQ, TQ test sets, and MS-pas development set. The results of the baselines are from original papers. \textbf{AR2+passage sive only runs with half of the batch size as AR2 and AR2+SimANS in NQ and TQ datasets, while only with 1/4 batch size as AR2+SimANS in MS-pas dataset}.}
\label{tab:mainexp}
\end{table*}
% Please add the following required packages to your passage preamble:
% \usepackage{booktabs}
% \usepackage{multirow}
\begin{table}[t]
\centering

\resizebox{0.95\columnwidth}{!}{%
\begin{tabular}{@{}c|lc|ccc@{}}
\toprule
\multicolumn{1}{l|}{} & \multicolumn{1}{c}{\textbf{Method}} & \multicolumn{1}{c|}{\textbf{B}} & \multicolumn{1}{c}{\textbf{R@5}} & \multicolumn{1}{c}{\textbf{R@20}} & \multicolumn{1}{c}{\textbf{R@100}} \\ \midrule
\multirow{4}{*}{NQ}   & AR2                                 & 64                                       & 77.9                               & 86.0                                & 90.1                                 \\
                      & AR2+SimANS                          & 64                                       & 78.6                               & 86.2                       & 90.3                                 \\
                      & AR2+passage sieve                   & 32                                       & 78.6                      & 86.1                                & 90.5                        \\
                      & AR2+passage sieve                   & 64                                      & \textbf{79.2}                      & \textbf{86.4}                                & \textbf{90.6}                        \\\midrule
\multirow{4}{*}{TQ}   & AR2                                 & 64                                       & 78.2                               & 84.4                                & 87.9                                 \\
                      & AR2+SimANS                          & 64                                       & 78.6                               & 84.6                                & 88.1                                 \\
                      & AR2+passage sieve                   & 32                                       & 78.6                      & \textbf{84.8}                       & \textbf{88.3}                        \\
                      & AR2+passage sieve                   & 64                                    & \textbf{78.7}                      & 84.7                                & 88.2                       \\\bottomrule
\end{tabular}
}
\caption{Comparison of AR2, AR2+SimANS and AR2+passage sieve, where \textbf{B} stands for batch size. Here, the number of hard negatives is set to 15 which is the same as in Section \ref{sec:passage-sieve-results}.}
\label{tab:ar2_simans}
\end{table}

% \begin{figure}[t]
%     \centering
%     \subfigure[NQ dataset]{\includegraphics[width=0.48\columnwidth]{fig/NQ-simans1.png}}
%     \subfigure[TQ dataset]{\includegraphics[width=0.48\columnwidth]{fig/TQ-simans1.png}}
%     \caption{Comparson of AR2, AR2+SimANS and AR2+passage sieve; Our methods achieve comparable performance on NQ and better performance on TQ with only half of the batch size}
%     \label{fig:simans}
% \end{figure}

\subsection{Experimental Setup}

\paragraph{Datasets} We conduct experiments on three public QA datasets: Natural Question (NQ) \cite{dataset-NQ}, Trivia QA (TQ) \cite{dataset-triviaqa} and MSMARCO Passage Ranking (MS-pas) \cite{dataset-ms}. The detailed statistics are given in the supplementary material.

\paragraph{Evaluation Metrics}
Following previous works, we report  R@k (k=5, 20, 100) for NQ and TQ, and MRR@10, R@k (k=50, 1K) for MS-pas. Here, MRR refers to the Mean Reciprocal Rank that calculates the reciprocal rank where the first relevant passage is achieved, and R@k measures the proportion of relevant passages (recall) in top-k results.

% The details of datasets, baselines, and implementations are presented in Appendix.
% We use AR2 \cite{AR2} as the base model to test the performance of the passage sieve algorithm \ref{algo:passage_sieve}. AR2 is a recent SOTA dense passage retrieval method that leverages the adversarial retriever-ranker and contrastive minimax objective instead of simple contrastive loss. We use the fundamental method in the dense retrieval field, DPR \cite{dpr}, as the base model to test the performance of our proposed robust contrastive loss. DPR leverages a bi-encoder architecture to encode the passage and query and NCE loss to train the model. Most modern dense retrieval methods originate from DPR and so using DPR to evaluate our approach is representative.
% \usepackage{graphicx}

\subsection{Effects of Passage Sieve Method}\label{sec:passage-sieve-results}
\subsubsection{Experimental Design}
We implement our passage sieve method on top of AR2 \cite{AR2}, which is referred to as AR2+passage sieve.  To evaluate the effectiveness of our passage sieve method, we compare it with the original AR2 along with other contemporary baselines. The details on the baselines are given in the supplementary material. 

During the passage sieve procedure in AR2+passage sieve, we set $\beta=0.5$ and epochs $T=1$, learning rate $\alpha=1e-7$. Additionally, we leverage the cosine similarity function and set the initial number of hard negative passages to be two times the number of hard negatives to be used in the downstream AR2 model. As for AR2 training, all settings are set the same way as in \cite{AR2}. Specifically, the batch size is set to 64 and the number of hard negatives is 15.
\subsubsection{Overall Results}

\begin{table}[t]

\centering

\resizebox{0.95\columnwidth}{!}{%
\begin{tabular}{l|l|ccc}
\toprule
\multicolumn{1}{c|}{\multirow{2}{*}{\textbf{Method}}} & \multicolumn{1}{c|}{\multirow{2}{*}{\textbf{\# hn}}} & \multicolumn{3}{c}{\textbf{NQ}} \\
\multicolumn{1}{c|}{} & \multicolumn{1}{c|}{} & \textbf{R@5} & \textbf{R@20} & \textbf{R@100} \\ \midrule
AR2 & 1 & 76.4 & 85.3 & 89.7 \\
AR2+passage sieve & 1 & \textbf{77.6} & \textbf{86.1} & \textbf{90.3} \\ \hline
AR2 & 5 & 76.9 & 85.3 & 89.7 \\
AR2+passage sieve & 5 & \textbf{78.0} & \textbf{85.9} & \textbf{90.6} \\ \hline
AR2 & 15 & 77.9 & 86.0 & 90.1 \\
AR2+passage sieve & 15 & \textbf{78.6} & \textbf{86.1} & \textbf{90.5} \\ \bottomrule
\end{tabular}
}
\caption{Performance of AR2 + passage sieve with the robust contrastive loss on NQ test sets. Here, \# hn stands for the number of hard negatives during training. We set the batch size to 32 for all compared methods. The results of the baselines are from \cite{AR2}.}
\label{tab:ablation}
\end{table}

Table \ref{tab:mainexp} presents the results of AR2+passage sieve on NQ and TQ test sets, as well as  MS-pas development set. It can be observed that AR2 outperforms most of the baseline models on all three datasets across different evaluation metrics. Furthermore, the proposed passage sieve approach enhances the performance of AR2 on all three datasets and evaluation metrics. The findings suggest that the sieve algorithm significantly contributes to improving the quality of hard negative samples in the dataset, enabling better results even with less GPU memory requirement. It is worth mentioning that our passage sieve procedure only accounts for one-tenth of the training time compared to AR2, the downstream retrieval model.

SimANS \cite{simans} is the recently proposed method to address the issue of false negatives based on some useful heuristic assumptions. When combined with AR2, AR2+SimANS achieves state-of-the-art results in dense retrieval. In contrast, we start from theory in noise-robust machine learning algorithms and develop a new loss function. The results in Table \ref{tab:ar2_simans} show that both SimANS and our method enhance the performance of AR2. However, compared to AR2+SimANS, the passage sieve achieves better results in all terms of TQ and NQ datasets. The advantage is more clearly seen on NQ dataset. It should be noted that we do not include the result on the MS-pas dataset of SimANS because they leverage a 4 times larger batch size than AR2 which makes the comparison unfair.

% \begin{table}[t]
% \centering
% \begin{tabular}{@{}lcccc@{}}
% \toprule
% Beta & \textbf{R@1} & \textbf{R@ 5} & \textbf{R@ 20} & \textbf{R@100} \\ \midrule
% 0    & 0.585        & 0.783         & 0.865          & 0.907          \\
% 0.25 & 0.578        & 0.784         & 0.866          & 0.906          \\
% 0.5  & 0.605        & 0.792         & 0.864          & 0.907          \\
% 1    & 0.601        & 0.791         & 0.867          & 0.906          \\ \bottomrule
% \end{tabular}

% \caption{Performance of AR2+passage sieve with different hyper-parameter $\beta$ on NQ test sets. Overall, the performance is not sensitive to $\beta$ in Recall@20 and Recall@100, however, if $\beta$ is too small, the R@1 and R@5 drop significantly. }
% \label{tab:beta}
% \end{table}
\begin{figure}[t]
    \centering
    \includegraphics[width=0.65\columnwidth]{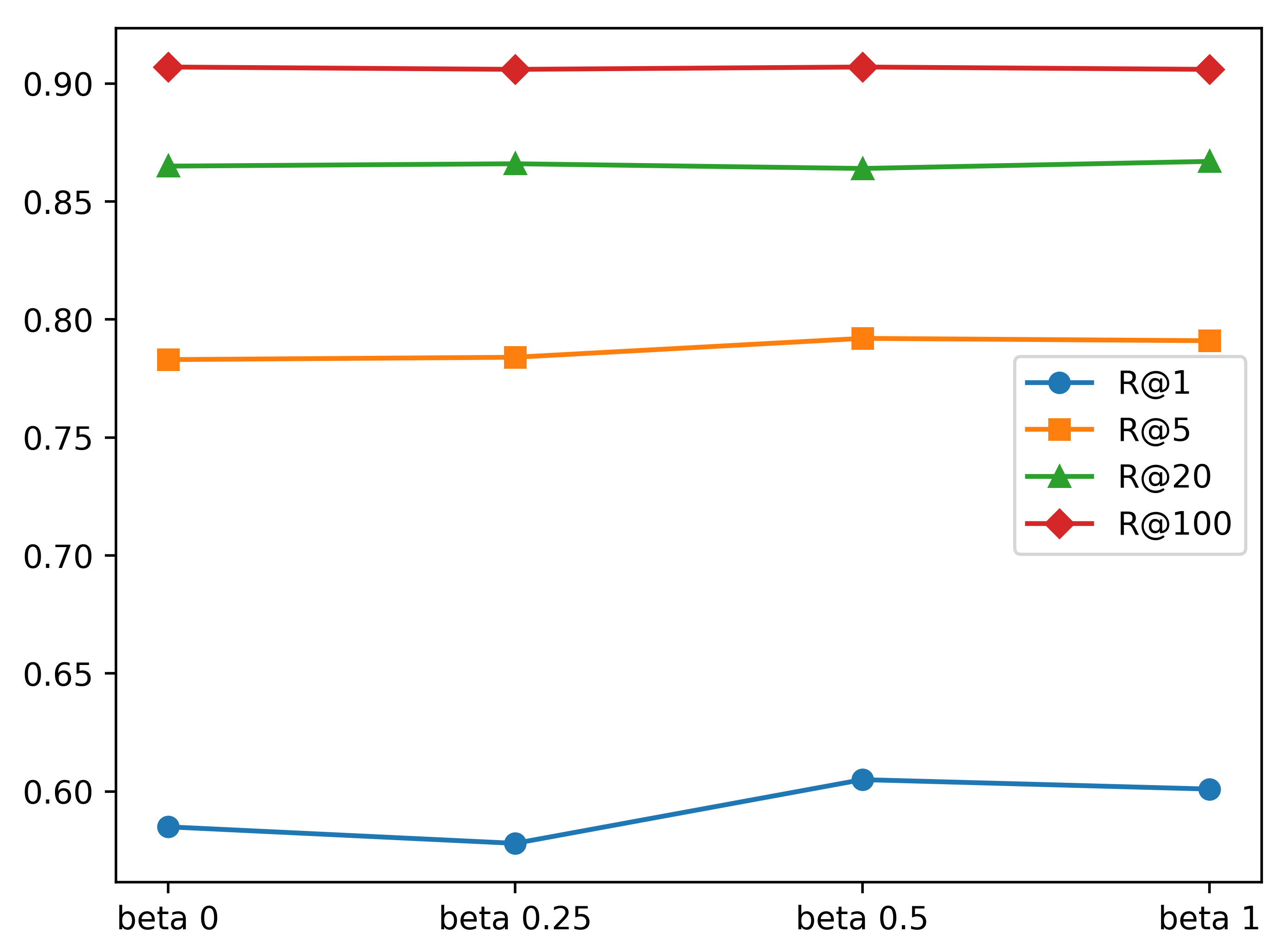}
    \caption{Performance of AR2+passage sieve with different hyper-parameter $\beta$ on NQ test sets. Overall, the passage sieve algorithm is not sensitive to $\beta$.}
    \label{fig:beta}
\end{figure}

\subsection{Detailed Analysis}

\paragraph{Influence of limited batch size}

We investigate the impact of smaller batch size on the performance of AR2+passage sieve. Table \ref{tab:ar2_simans} shows that AR2+passage sieve achieves better results in terms of R@5, R@20, and R@100 on TQ dataset, even with a smaller batch size. On NQ dataset, our method achieves comparable results with AR2+SimANS while having only a slight decrease in R@20. 
\paragraph{Influence of the number of hard negatives}
To evaluate the impact of hard negatives, we experiment with varying numbers of hard negatives during training. To speed up the experiment, the batch size is set to 32 and we only work on NQ dataset. The results in Table \ref{tab:ablation} demonstrate that the passage sieve approach consistently enhances the performance of AR2 across all settings. Particularly, with only 5 hard negatives, our method can reach a performance near that of 15 hard negatives. Intuitively, when the quality of hard negative samples is higher, we do not need many negative samples, thus reducing computing resources for training. 
%Particularly noteworthy is the substantial improvement of AR2 achieved by the passage sieve technique, especially in cases where there is a scarcity of hard negative passages. %This highlights the effectiveness of the passage sieve methodology in carefully curating the dataset and effectively removing false negative cases, consequently reducing the need for a large number of hard negative passages in retrieval algorithm training.

\paragraph{Influence of hyper-parameter $\beta$}
To study the influence of the hyper-parameter $\beta$ on the passage sieve algorithm, we conduct experiments on NQ dataset with different $\beta$ and all other settings remain the same. Specifically, we vary the values of $\beta$ to be 0, 0.25, 0.5. Figure \ref{fig:beta} indicates the passage sieve method is not sensitive to $\beta$. This is because DPR in the passage sieve method is trained with cosine similarity, which satisfies our assumptions in Theorem 2.

\begin{table}[t]
\centering
\resizebox{0.9\columnwidth}{!}{%
\begin{tabular}{c|ll|ccc}
\toprule
\multicolumn{1}{l|}{} & \multicolumn{1}{c}{\textbf{Method}} & \multicolumn{1}{c|}{\textbf{Batch size}} & \textbf{\textbf{R@5}} & \textbf{\textbf{R@20}} & \textbf{\textbf{R@100}} \\ \midrule
\multirow{3}{*}{NQ} & DPR & 128 & - & 78.4 & 85.4 \\
 & DPR+CCR & 64 & 65.9 & 77.6 & 85.4 \\
 & DPR+CCR & 128 & \textbf{68.4}&\textbf{79.5} & \textbf{86.1} \\ \hline
\multirow{3}{*}{TQ} & DPR & 128 & - & 79.3 & 84.9 \\
 & DPR+CCR & 64 &70.5  & 78.7 & 84.9  \\
 & DPR+CCR & 128 & \textbf{71.5} & \textbf{79.8} &  \textbf{85.1} \\ \bottomrule
\end{tabular}%
}
\caption{Performance of DPR+Contrastive Confidence Regularizer on NQ test sets. The results of the baselines are from original papers.}
\label{tab:dprres}
\end{table}

\subsection{Effects of Contrastive Confidence Regularizer}
\subsubsection{Exterimental Design}
We evaluate the performance of the Contrastive Confidence Regularizer on the DPR model \cite{dpr}. We incorporate the regularizer into the existing NCE loss function during the final 5 epochs of training. Other parameters and configurations are kept consistent with those in the original DPR paper. Since the DPR model utilizes a dot-product similarity function, we select the values of $\beta$ from the range of 0.001 to 0.0001. The selection of $\beta$ is based on the performance observed on the validation set.

\subsubsection{Experiment Results}
Table \ref{tab:dprres} shows results on NQ and TQ datasets. It is observable that the proposed contrastive confidence regularizer helps DPR get a better performance across all metrics on NQ and TQ datasets. The experimental results also show that the reduction of batch size has a greater impact on R@20 and R@5, but a smaller impact on R@100. On both datasets, the proposed CCR helps achieve the same R@100 value as DPR with only half the batch size. 

\section{Conclusion}
This paper aims to mitigate the impact of false negatives on dense passage retrieval. Toward such a goal, we extend the peer-loss framework and develop a confidence regularization for training robust retrieval models. The proposed regularization is compatible with any base retrieval model that uses NCE loss, a widely used contrastive loss function in dense retrieval. Through empirical and theoretical analysis, it is demonstrated that contrastive confidence regularization leads to more robust retrieval models. Building on this regularization, a passage sieve algorithm is proposed. The algorithm leverages a dense retrieval model trained with confidence regularized NCE loss to filter out false negatives, thereby improving any downstream retrieval model including those that do not exploit NCE loss. The effectiveness of both the passage sieve algorithm and the confidence regularization method is validated through extensive experiments on three commonly used QA datasets. The results show that these methods can enhance base retrieval models, even when fewer negative samples are used.

% \section{Acknowledgments}
% AAAI is especially grateful to Peter Patel Schneider for his work in implementing the original aaai.sty file, liberally using the ideas of other style hackers, including Barbara Beeton. We also acknowledge with thanks the work of George Ferguson for his guide to using the style and BibTeX files --- which has been incorporated into this passage --- and Hans Guesgen, who provided several timely modifications, as well as the many others who have, from time to time, sent in suggestions on improvements to the AAAI style. We are especially grateful to Francisco Cruz, Marc Pujol-Gonzalez, and Mico Loretan for the improvements to the Bib\TeX{} and \LaTeX{} files made in 2020.

% The preparation of the \LaTeX{} and Bib\TeX{} files that implement these instructions was supported by Schlumberger Palo Alto Research, AT\&T Bell Laboratories, Morgan Kaufmann Publishers, The Live Oak Press, LLC, and AAAI Press. Bibliography style changes were added by Sunil Issar. \verb+\+pubnote was added by J. Scott Penberthy. George Ferguson added support for printing the AAAI copyright slug. Additional changes to aaai24.sty and aaai24.bst have been made by Francisco Cruz and Marc Pujol-Gonzalez.

% \bigskip
% \noindent Thank you for reading these instructions carefully. We look forward to receiving your electronic files!
\section*{Acknowledgments}
We thank all the reviewers for their helpful comments. This work is supported by the National Key R\&D Program of China (2022ZD0116600).
\bibliography{aaai24}

\newpage
\appendix
\section*{Appendix}
\section{Missing Details}
\subsection{Limitations and future work}
This paper proposed a novel contrastive confidence regularizer and a passage sieve algorithm to handle the false negative problem in dense retrieval training. The theory analysis and experimental results have shown the effectiveness of our methods, however, there are still several limitations of our work:
\begin{enumerate}
    \item We only test our methods on top of AR2 \cite{AR2} and DPR \cite{dpr}, more retrieval models can be tested to see the performance variance of the passage sieve and contrastive confidence regularizer in different cases.
    \item Theorem 1  only assumes the loss function to be NCE loss and does not require the task to be dense retrieval. Thus, the proposed contrastive confidence regularizer has the same power on any other self-supervised tasks using NCE loss, e.g. self-supervised image classification, sentence embedding, etc. More experiments on these benchmarks should be conducted in future work.
    \item The proposed methods are built upon $\text{CORES}^2$ \cite{cores2}, which consider the first-order statistic(estimation) of peer loss. However, \citet{second-order} has extended this into considering the second-order statistics and achieving better performance in CV benchmarks. More work is needed to extend the contrastive confidence regularizer into second-order statistics as well.
\end{enumerate}
\subsection{Details about baselines}
Details about the baselines in Table 1 in the main content are shown in this subsection.\\
\textbf{DPR} \cite{dpr} employs a dual-encoder architecture, leveraging BM25 for the identification of hard negatives and utilizing the in-batch negatives for efficiency. \\
\textbf{ANCE} \cite{ance} constructs hard negatives from an Approximate Nearest Neighbor (ANN) index. \\
\textbf{COIL} \cite{coil} integrates semantic with lexical matching, storing contextualized token representations in inverted lists.\\
\textbf{ME-BERT} \cite{me-bert} merges the efficiency of dual encoders with the expressiveness of attentional architectures, further venturing into sparse-dense hybrids. \\
\textbf{Individual top-k} \cite{individual-topk} utilizes unsupervised pre-training, focusing on the inverse cloze task and masked salient spans within question-context pairs.\\
\textbf{RocketQA} \cite{rocketQA} explores the ability of the cross-encoder model to denoise hard negatives. \\
\textbf{RDR} \cite{RDR}  seeks to distill knowledge from the reader into the retriever, aiming for the retriever to gain reader strengths without loss of inherent advantages.\\ %proposes to distill the reader into the retriever so that the retriever absorbs the strength of the reader while keeping its own benefit\\
\textbf{RocketQAv2} \cite{rocketqav2} distills knowledge from the cross-encoder model by narrowing the relevance distribution between the retrieval model and the cross-encoder model.  \\
\textbf{PAIR} \cite{Pair} considers both query-centric similarity relations and passage-centric similarity relations.\\
\textbf{DPR-PAQ} \cite{DPR-PAQ} introduces the PAQ dataset, concentrating on the question-answering pretraining task.\\
\textbf{Condenser} \cite{condenser} boosts optimization readiness by performing mask language model predictions actively condition on dense representation.\\
\textbf{coCondenser} \cite{cocondenser} adds a corpus-level contrastive learning goal to the existing Condenser pre-training architecture.\\
\textbf{ERNIE-Search} \cite{ernie} refines the distillation process by sharing the encoding layers with the dual-encoder and conducting a cascaded approach.\\
\textbf{MVR} \cite{mvr} offers multiple representations for candidate passages and proposes a global-local loss to prevent multiple embeddings from collapsing to the same. \\
\textbf{PROD} \cite{lin2023prod} enhances the cross-encoder distillation by cyclically improving teacher models and tailoring sample selections to student needs.\\ %improves the distillation from cross-encoder by gradually improving the capability of teachers by using different architectures enabling students to learn knowledge progressively and gradually select samples that the student is confused about to suit the requirement of the student at different stages. \\
\textbf{COT-MAE} \cite{cot-mae} utilizes an asymmetric encoder-decoder architecture with self-supervised and context-supervised masked auto-encoding tasks.\\%employs an asymmetric encoder-decoder architecture that learns to compress the sentence semantics into a dense vector through self-supervised and context-supervised masked auto-encoding.\\
\textbf{AR2} \cite{AR2} jointly optimizes the retrieval and re-ranker model according to a minimax adversarial objective.

\subsection{Example implementation of contrastive confidence regularizer}
In section 5, we mentioned that CCR (eq.8 in the main content) only slightly increases the time complexity when implemented properly. In this subsection, we will demonstrate the implementation to prove this claim.

The example code is shown in listing \ref{lst:python}. The dimension of $s$ is the number of queries in the batch times the number of passages (including positive, in-batch negatives, and hard negatives) of each query. In order to calculate the NCE loss, a log softmax function on the passage dimension is required. And after that, an NLL loss with only positive passages labeled as 1 is calculated with the after-softmax matrix $s\_$. The only additional code for CCR is to calculate the NLL loss for all elements in the $s\_$ and get the mean value of each row of it.
\begin{lstlisting}[language=Python, caption= NCE loss with Contrastive Confidence Regularizer, label=lst:python, frame=single, escapeinside={(*@}{@*)}]
s = get_scores(q, p) #get the similarity matrix
s_ = F.log_softmax(s, dim=1)#calculate softmax on passage dimension
loss = F.nll_loss(s_,positive_idx,reduction="none") #the original NCE loss
loss=loss-beta*(-1)*s_.mean(dim=1)#the only additional code to calculate CCR
loss=loss.mean()
\end{lstlisting}

\subsection{Efficiency analysis of passage sieve algorithm}
Overall speaking, The time taken to integrate the passage sieve (algorithm 1) into AR2 is equivalent to that of training a vanilla DPR\cite{dpr} model. Compared with the time consumption of AR2 training itself, the additional training time is relatively small. We record the time consumption of one epoch training on the TQ dataset and show it in table \ref{tab:efficiency}.

It should be noted that the passage sieve has nothing to do with the inference time consumption. Adding the passage sieve only requires an additional data preprocessing time during training.
% Please add the following required packages to your document preamble:
% \usepackage{booktabs}
\begin{table}[ht]
\centering
\resizebox{0.95\columnwidth}{!}{%
\begin{tabular}{@{}l|cc@{}}
\toprule
                                & AR2         & AR2+passage sieve \\ \midrule
Batch size                      & 32          & 32                \\
Number of hard negatives        & 15          & 15                \\
Data preprocessing              & 10min       & 2h 10min                \\
Retrieval \& Reranker training & 2h 5min     & 2h 5min           \\
Corpus and query encoding       & 6h 6min     & 6h 6min           \\
ANN index build and Retrieval   & 30min       & 30min             \\\midrule
\textbf{Overall}                & \textbf{9h} & \textbf{11h}      \\ \bottomrule
\end{tabular}
}
\caption{Comparison of efficiency of AR2 + passage sieve  and AR2 on the TQ dataset. We list the time consumption for one epoch of training with 4 Tesla V100s.}
\label{tab:efficiency}
\end{table}
\subsection{Intermediate experiment result}
\paragraph{Sieve out rate of passage sieve} In the passage sieve algorithm, we perform passage sieves according to their loss value and the average loss value (i.e. threshold). And we prove that those who get a score $f(p,q)$ more than the average value will be sieved out for sure, however, it remains unknown what percentage of all negative passages will be sieved out (we called it sieve out rate). The sieve out rate directly affects how many passages we need to examine during the passage sieve in order to meet the hard negatives requirement of the downstream retrieval model. 

\begin{figure}[ht]
    \centering
    \includegraphics[width=0.95\columnwidth]{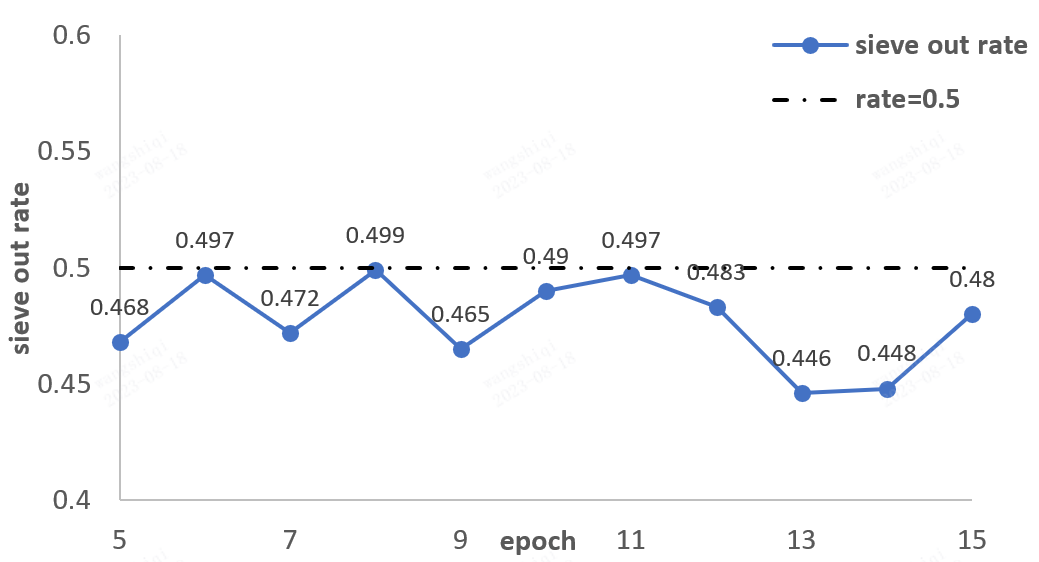}
    \caption{Sieve out rate during AR2+passage sieve training on NQ dataset}
    \label{fig:sieverate}
\end{figure}

In the main content, we mentioned that we choose two times the number of hard negatives that the downstream retrieval model (AR2 in this case) will use. This ratio is actually chosen by some preliminary experiments, in which we observed that the passage sieve algorithm usually sieves out examples less than half. We show the sieve-out rate during training on the NQ dataset in figure \ref{fig:sieverate}. The figure shows that the sieve rate is usually smaller than but close to 0.5 and usually when one epoch achieves a sieve-out rate near 0.5, the rate of the next epoch will drop immediately. It should be noted that this rate has nothing to do with the estimated noise pair possibility. The strategy adopted by the passage sieve algorithm is to filter aggressively and only retain confident ones, in which only the sufficient conditions for filtering are specified (see Lemma 1).  

\begin{figure}[ht]
    \centering
    \subfigure[TQ test set]{\includegraphics[width=0.48\columnwidth]{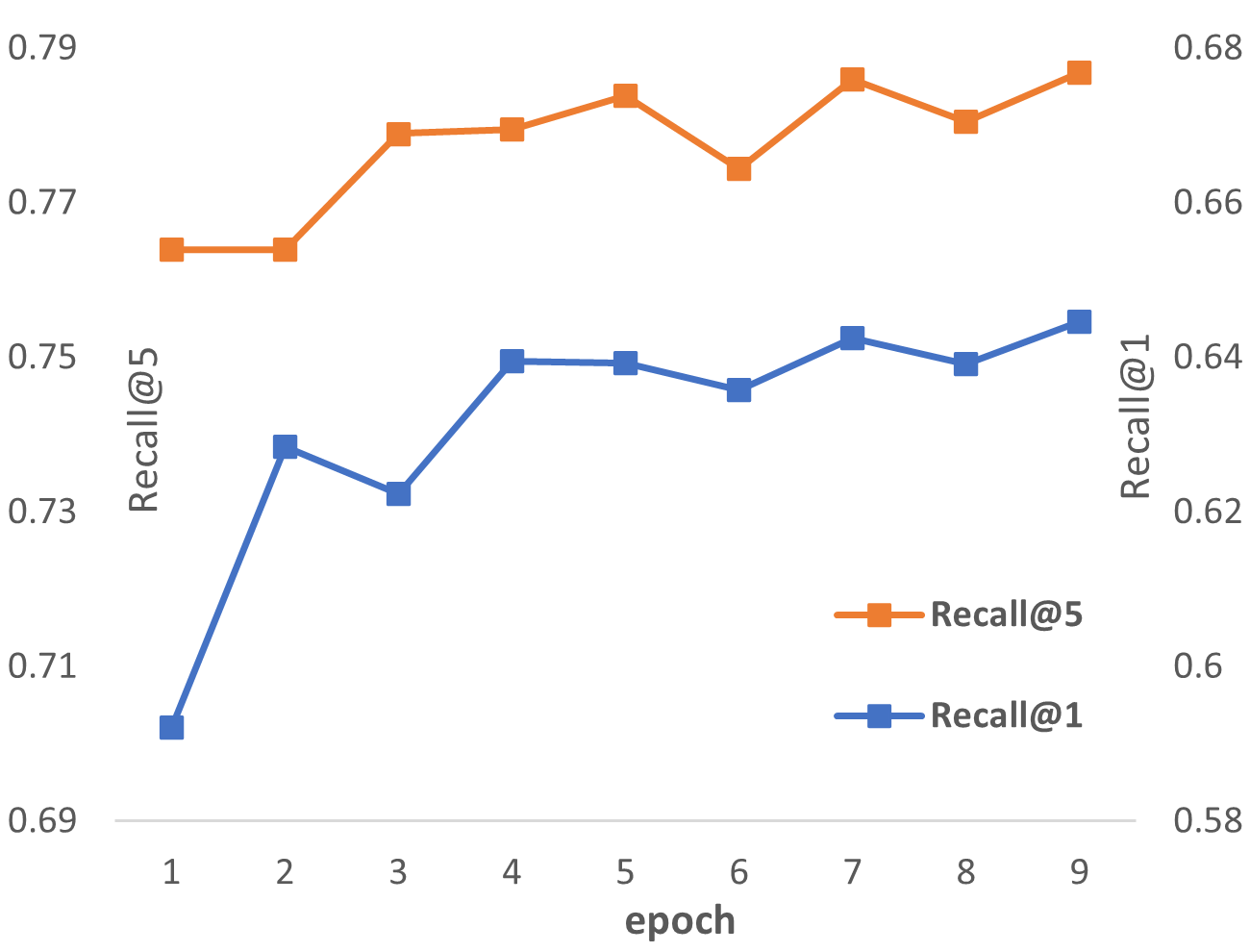}}
    \subfigure[MS-pas dev set]{\includegraphics[width=0.48\columnwidth]{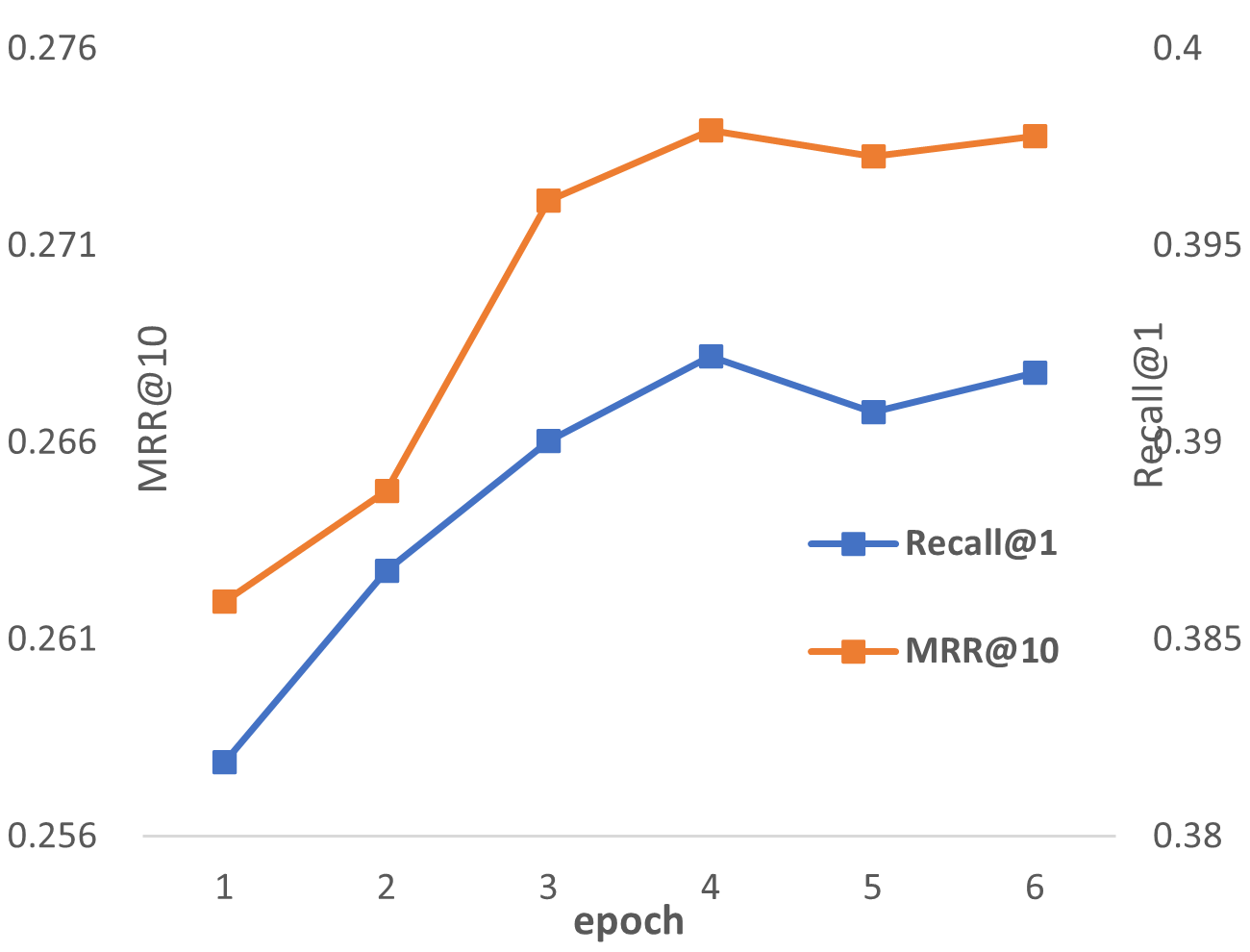}}
    \caption{Learning curve of MS-pas development set and TQ test set. }
    \label{fig:curve}
\end{figure}

\paragraph{Learning curves}
We also show the visualization of the training procedure. Specifically, the Recall@1 Recall@5 of the TQ test set and Recall@1 MRR@10 of the MS-Pas dataset are shown in the figure. And we can see from fig.\ref{fig:curve} that the metrics on both TQ and MS-pas datasets gradually increase until convergence.
\subsection{Annotation Table}
We conclude the annotation we used in the main content of the paper and show them in table \ref{tab:annotation} to help read our paper.
% Please add the following required packages to your document preamble:
% \usepackage{booktabs}
\begin{table}[ht]
\resizebox{\columnwidth}{!}{%
\begin{tabular}{@{}ll|ll@{}}
\toprule
$p$ &passage  & $\mathcal{D}$  & clean distribution  \\
$q$&query  & $\mathcal{D}_{P|q}$ & distribution of P given q   \\
$p^-$& negatives passage  &$\ell_{CCR}$  & contrastive confidence regularizer  \\
$p^+$&positive passage  & $\ell_{RCL}$  & robust contrastive loss  \\
$\mathcal{N}_{q}$ & negative set for query q  & $\mathcal{D}^*$ & sieved dataset  \\
$\tilde{\mathcal{D}}$ & Noisy distribution  & $\mathcal{N}_{q}^*$ & confident negative list \\
$\ell_{NCE}$ & NCE loss &  $f(p,q)$ & similarity score of p,q after softmax    \\
\bottomrule
\end{tabular}
}
\caption{Annotation table}
\label{tab:annotation}
\end{table}
\section{Important Proofs}
Before we start the proof, we will first introduce the pseudo-label and corresponding loss function we used during the proof. Let's consider a positive pair $(p^+,q)$ as label $y=1$ and negative pair $(p^-,q)$ as label $y=-1$. Remind that $p$ means passages and $q$ stands for queries. In the following context, instead of using superscript, we will leverage this label to indicate the relation between passages and queries. And similar to the context in the label-noise robust machine learning papers, $y$ represents the real label and $\tilde{y}$ for the noisy label in the dataset. Thus we can represent the false negatives in the dataset as $(p,q;\tilde{y}=-1,y=1)$.  We then re-define the contrastive loss function as :
\begin{equation}
\ell(f(p,q);y)=
    \begin{cases}
    \ell_{NCE}(p,q), & y=1,\\
    0, & y=-1,
    \end{cases}
    \label{eq:pseudo-loss}
\end{equation}
where 
\begin{equation}
\begin{aligned}
    \ell_{NCE}(p,q)&=-\ln(f(p,q))\\
    &=-\ln(\frac{e^{sim(p,q)}}{\sum_{p_i^- \in \mathcal{N}_{q}}e^{sim(p_i^-,q)}+e^{sim(p^+,q)}})
\end{aligned}
\end{equation}
Here, $f$ is the model that gives the score between queries and passages. $\ell_{NCE}$ is the broadened NCE loss as we discussed in the preliminary part. This newly defined pseudo loss is for proof purposes only which helps better analyze the performance of the proposed methods during proof. Pay attention that such a definition will not affect the original contrastive learning in dense retrieval as the loss for negative pairs is always equal to 0. And there is only a coefficient difference between them:
\begin{equation*}
    \scalebox{0.9}{
       $\mathbb{E}_{(P,Q;Y)\in D}\left[\ell(f(P,Q);Y)\right] = \mathbb{P}(Y=1) \mathbb{E}_{(P^+,Q)\in D}\left[\ell_{NCE}(P,Q)\right]$
        }
\end{equation*}
In the following context, unless otherwise specified, $\ell$ stands for this defined loss, and $\ell_{NCE}$ stands for the original NCE loss.

\subsection{Proof for theorem 1 and theorem 2}

To make the reading more natural and smooth, we simultaneously prove Theorems 1 and 2 in this section.\\
\textbf{Theorem 1. }
\textit{
    With the assumption that the possibility of a noisy pair is smaller than a clean pair and a suitable selection of $\beta$, we have: minimizing $\mathbb{E}_{\tilde{\mathcal{D}}}[\ell(p^+,q)-\beta*\ell_{CCR}(p^+,q)]$ is equivalent to minimizing $\mathbb{E}_\mathcal{D}[\ell(p^+,q)]$}\\
\textbf{Theorem 2. }
\textit{
    When the assumption in Theorem 1 holds, the similarity function $sim$ is bounded (e.g. cosine similarity)  and there is no false positive in the dataset $\tilde{\mathcal{D}}$, the $\beta$ that satisfies Theorem 1 must exist and in the interval [0,1]. }
\begin{proof}
    We first prove that with the new regularizer, the Main Theorem of \cite{cores2}, which decouples the expected regularized loss, still holds. For simplicity, considering $X:=(p_n,q_n)$

    We define the new pseudo loss as in eq.\ref{eq:pseudo-loss}, and this loss is aligned with the basic form of the classification loss function with feature x and label y which is used by $\text{CORES}^2$ \cite{cores2}. And we first get those that do NOT require the specific form of the loss function and thus remain unchanged in our case as the original confidence regularizer paper as the eq.\ref{eq:same_eq} follows:
    \begin{equation}
        \begin{aligned}
            &\mathbb{E}_{\tilde{\mathcal{D}}}[\ell(f(X),\tilde{Y})]=
            \sum_{j}T_{jj}\mathbb{P}(Y=j)\mathbb{E}_{\mathcal{D}|Y=j}\left[\ell(f(X),j)\right]+\\&\sum_{j}\sum_{\substack{i}}\mathbb{P}(Y=i)\mathbb{E}_{\mathcal{D}|Y=i}[U_{ij}(X)\ell(f(X),j)]\\    
            &=\underline{T}\mathbb{E}_{\mathcal{D}}[\ell(f(X),Y)]+\bar{\Delta}\mathbb{E}_{\mathcal{D}_{\Delta}}[\ell(f(X),Y)]+\\&\sum_{j\in\{-1,1\}}\sum_{\substack{i\in\{-1,1\}}}\mathbb{P}(Y=i)\mathbb{E}_{\mathcal{D}|Y=i}[U_{ij}(X)\ell(f(X),j)],
        \end{aligned}
        \label{eq:same_eq}
    \end{equation}
    where
    \begin{equation*}
        \begin{aligned}
            &T_{ij}:=\mathbb{E}_{\mathcal{D}|Y=i}[T_{ij}(X)],  \forall i,j\in[-1,1]\\
            &T_{ij}(X):=\mathbb{P}(\tilde{Y}=j|Y=i,X), \underline{T}=\min\{T_{-1-1},T_{11}\}\\
            % &\Delta_j:=T_{jj}-\underline{T}, \bar{\Delta}:=\sum_{j\in\{-1,1\}}\Delta_j \mathbb{P}(Y=j)\\
            &U_{ij}(X)=T_{ij}(X),\forall i\neq j,\quad U_{jj}(X)=T_{jj}(X)-T_{jj}
            \end{aligned}
    \end{equation*}
    % \begin{equation*}
    %     \begin{aligned}
    %         &\mathbb{E}_{\mathcal{D}_{\Delta}}[\ell(f(X),Y)]:=\mathbbm{1}(\bar{\Delta}\geq 0) \times 
    %         \\ &\sum_{j\in\{-1,1\}}\frac{\Delta_j\mathbb{P}(Y=j)}{\bar{\Delta}}\mathbb{E}_{D|Y=j}[\ell(f(X),j)].
    %     \end{aligned}
    % \end{equation*}
     $\bar{\Delta}$ and $\mathcal{D}_\Delta$ follow the same definition in $\text{CORES}^2$, and we don't list them here as it's not required in our proof. 
    Furthermore, as $\ell(f(X); Y=-1)=0$ in the pseudo loss (eq. \ref{eq:pseudo-loss}) we defined, we can further get the new decoupled NCE loss in our problem:
    \begin{equation}
        \begin{aligned}
            &\mathbb{E}_{\tilde{\mathcal{D}}}[\ell(f(X),\tilde{Y})]=\\ &
            \underline{T}\mathbb{E}_{\mathcal{D}}[\ell(f(X),Y)]+\bar{\Delta}\mathbb{E}_{\mathcal{D}_{\Delta}}[\ell(f(X),Y)]+\\&\sum_{\substack{i\in\{-1,1\}}}\mathbb{P}(Y=i)\mathbb{E}_{\mathcal{D}|Y=i}[U_{i,1}(X)\ell(f(X),1)]
        \end{aligned}
        \label{eq:decoulped_NCE}
    \end{equation}
     We use a novel contrastive confidence regularizer which is different from the original regularizer in $\text{CORES}^2$. And the expected form of the new $\ell_{CCR}(p_n,q_n)$ is
    \begin{equation}
        \begin{aligned}
            \mathbb{E}_{\tilde{\mathcal{D}}}\left[\ell_{CCR}(p_n,q_n)\right]&=\mathbb{E}_{\mathcal{D}}\left[\ell_{CCR}(p_n,q_n)\right]\\
            &=\mathbb{E}_{q_n}\left[\mathbb{E}_{\mathcal{D}_{p|q_n}}[\ell(f(p,q_n);1)]\right]\\
            &=\mathbb{E}_{\mathcal{D}_{p,q}}[\ell(p,q;1)]\\
            &=\sum_{i\in\{-1,1\}}\mathbb{P}(Y=i)\mathbb{E}_{\mathcal{D}|Y=i}[\ell(p,q;1)]
        \end{aligned}
        \label{eq:ccr_est}
    \end{equation}
    The first equation in eq.\ref{eq:ccr_est} is true because the clean distribution and the noisy distribution differ from labels instead of distributions of $p$ and $q$.  It helps bridge the clean distribution and the noisy distribution. And with the final equivalent transformation in eq.\ref{eq:ccr_est}, we are able to merge the third term of the decoupled NCE loss eq.\ref{eq:decoulped_NCE} and the expectation of CCR together. Thus the expectation of regularized loss is: 
    \begin{equation}
        \begin{aligned}
            &\mathbb{E}_{\mathcal{\tilde{D}}}\left[\ell(f(X),\tilde{Y})-\beta*\ell_{CCR}(X)\right]=\\
            &\underline{T}\mathbb{E}_{\mathcal{D}}[\ell(f(X),Y)]+\bar{\Delta}\mathbb{E}_{\mathcal{D}_{\Delta}}[\ell(f(X),Y)]+\\
            &\sum_{i\in\{-1,1\}}\mathbb{P}(Y=i)\mathbb{E}_{\mathcal{D}|Y=i}[(U_{i,1}(X)-\beta)\ell(f(X),1)]
        \end{aligned}
        \label{eq:e_rcl}
    \end{equation}
    
    After we get eq.\ref{eq:e_rcl}, the next key problem is to examine the influence of each term in it during training on noisy datasets. Optimizing the first term represents the optimization of the clean distribution and so we don't need to consider it. The second term represents expectation on a shifted distribution $\mathcal{D}_{\Delta}$. Lemma 2 in $\text{CORES}^2$ indicates that as long as the dataset is informative and the clean label $y$ equals the Bayes optimal label $y^*:= \arg\max_{y}\mathbb{P}(y|X)$, the second term won't change the Bayes optimal label. In our case, the second assumption is naturally satisfied as one sample (p,q) only belongs to one class (positive or negative) in the clean dataset. So, with the assumption that the noisy distribution is informative, i.e. \textbf{ ``Possibility of any pair to be noisy is smaller than clean (Assumption I)"}, the second term in eq.\ref{eq:e_rcl} does NOT change the Bayes optimal classifier during training. 

    Considering the third term in eq.\ref{eq:e_rcl}, the ideal condition is $U_{1,1}(x)=U_{-1,1}(x)=\beta, \forall x$ and so the third term will always be equal to zero. However, this condition is nearly impossible to meet and violates the assumption of instance-dependent noise. In the original confidence regularizer paper \cite{cores2}, they succeed in proving an interval of $\beta$ in which the regularized loss will produce the same Bayes optimal classifier as in the clean distribution.  Nevertheless, with a complex interval expression,  there is still a risk of nonexistence for beta due to the unclear relationship between the starting and ending points of the interval. 
    
    In contrast, the third term in eq.\ref{eq:e_rcl} is significantly different from the original one. And we have a more concrete problem setting. So with new assumptions associated with the problem, we can re-prove a new must-exist bound for $\beta$ to make the estimation unbiased for the clean datasets. Intuitively, the proof can be concluded as follows: Firstly, $U_{i,1}(X)$  is related to the relative clean rate of X compared to the average clean rate and so when $\beta$ is large enough, the scaling factor of the third term $U_{i,1}(X)-\beta$ will be negative, thus reversing the loss associated with instances with labeled noise. Secondly,  $\beta$ should not be too large, otherwise, it will affect the learning from clean data.
    
     On the one hand, when $(U_{i,1}(X)-\beta)\leq 0 \Leftarrow \beta \geq \max_{i\in\{-1,1\}, X \sim \mathcal{D}_x}U_{i,1}(X)$, all the scaling factor $(U_{i,1}(X)-\beta)$ would be negative, minimizing the third term equals to minimize the regularization term. It should be noted that, given the assumption that \textbf{``The NCE loss has an infimum, i.e. similarity function is bounded with a minimal value (Assumption II)"}, Theorem 1 in $\text{CORES}^2$\cite{cores2} still holds even when we use a different form of the loss function. According to this theorem, minimizing the regularization term will induce confidence prediction no matter whether the distribution is noisy or clean. 
    
    Pay attention that, given the assumption that \textbf{``There is no false positive problem in the distribution (Assumption III)"}, we have $T_{-1,-1}(X)=1, T_{-1,1}(X)=0, U_{-1,1}(X)=T_{-1,1}(X)=0$ in our problem and so the lower bound become:
    \begin{equation}
        \begin{aligned}
            \beta &\geq \max_{X \sim \mathcal{D}_x}U_{1,1}(X)=\max_{X \sim \mathcal{D}_x} T_{1,1}(X)-T_{1,1}\\
            &=\max_{X \sim \mathcal{D}_x} T_{1,1}(X)-\mathbb{E}_{\mathcal{D}|Y=1}[T_{1,1}(x)]
        \end{aligned}
    \end{equation}
    
    On the other hand, we still need to determine an upper bound on the beta to prevent the loss function from being biased against the clean data set. Consider a sample $x_n$ whose clean label is $y_n=1$, and the model $f(x)$ does not fit it well so that its loss $\ell(f(x_n),1)=\ell_0>0$. To make the regularized loss unbiased in this case, it should also be bigger than 0:

\begin{equation}
    \begin{aligned}
       & T_{1,1}\mathbb{P}(x_n)\ell_0 + \mathbb{P}(x_n)(U_{1,1}(x_n)-\beta)\ell_0 \geq 0\\
    &\Leftrightarrow T_{1,1} + (U_{1,1}(x_n)-\beta) \geq 0  \\ 
    & \Leftarrow \beta \leq T_{1,1} + \min_{X\sim \mathcal{D}_{X}}U_{1,1}(x)\\
    &\Leftrightarrow \beta \leq T_{1,1} + \min_{X\sim\mathcal{D}_{X}} (T_{1,1}(x)-T_{1,1}) \\
    &\Leftrightarrow \beta \leq \min_{X\sim \mathcal{D}_{X}}T_{1,1}(X) 
    \end{aligned}
\end{equation}

    And with the assumption that the noise rate is bounded as $T_{1,1}(X)>T_{1,-1}(X),\forall X \sim \mathcal{D}_X$, we will get that $T_{1,1}(x)\geq 1/2, \forall x$ and so 
    \begin{equation}
        \begin{aligned}
            &\max_{X \sim \mathcal{D}_x} T_{1,1}(X)-\mathbb{E}_{\mathcal{D}|Y=1}[T_{1,1}(x)]  \leq 1-1/2=1/2
        \end{aligned}
    \end{equation}
    And 
    \begin{equation}
        \min_{X\sim \mathcal{D}_{X}}T_{1,1}(X) >1/2
    \end{equation}
    so there must be feasible $\beta\in [0,1]$ which satisfies the two bounds at the same time and thus reduces unbiased confidence prediction of the clean dataset. And more concretely, $\beta=0.5$ is a safe choice all the time.
    
    In conclusion, with Assumptions I, II, and III, when
    $$
    \max_{X \sim \mathcal{D}_x} T_{1,1}(X)-T_{1,1}\leq \beta \leq \min_{X\sim \mathcal{D}_{X}}T_{1,1}(X)
    $$
    minimizing $\mathbb{E}_{\tilde{D}}[\ell(p,q;\tilde{Y})-\beta*\ell_{CCR}(p,q)]$ is equivalent to minimizing $\mathbb{E}_\mathcal{D}[\ell(p,q; Y)]$ in the dense retrieval problem and such $\beta$ must exist in [0,1].
\end{proof}

\subsection{Proof for Lemma 1}
\textbf{lemma 1.}
\textit{
The passage sieve algorithm (1) ensures that a negative sample $p^-$ in hard negatives will NOT be selected into the sieved dataset $\mathcal{D}^*$ if its score $f(p^-,q)$ given by the model f is more than a random guess, i.e. its similarity score after softmax is bigger than the average value $1/(|\mathcal{N}_k|+1)$}

\begin{proof} Considering the sieving process in (1). for a negative pair $(p_n,q_n;\tilde{y}_n=-1)$, NOT being selected into the confident negative set equals the following inequality, for simplicity, define $\mathcal{N}=\mathcal{N}_{q_n}\cup\{p^+_n\}$:
\begin{equation}
    \begin{aligned}
        &\ell(p_n,q_n;1) \leq \frac{1}{|\mathcal{N}_{q_n}|+1}\sum_{p_i \in \mathcal{N}}\ell(p_i,q_n;1)\\
        &\ell_{NCE}(p_n,q_n) \leq \frac{1}{|\mathcal{N}_{q_n}|+1}\sum_{p_i \in \mathcal{N}}\ell_{NCE}(p_i,q_n)\\
        \Leftrightarrow &\ln(f(p_n,q_n)) \geq \frac{1}{|\mathcal{N}_{q_n}|+1}\sum_{p_i \in \mathcal{N}}\ln(f(p_i,q_n))\\
        \Leftrightarrow &\ln(f(p_n,q_n)) \geq \frac{1}{|\mathcal{N}_{q_n}|}\sum_{p_i \in \mathcal{N}, p_i \neq p_n }\ln(f(p_i,q_n))
    \end{aligned}\tag{2-1}
\end{equation}
By Jensen's inequality, we have:
$$
\frac{1}{|\mathcal{N}_{q_n}|}\sum_{p_i \in \mathcal{N}, p_i \neq p_n}\ln(f(p_i,q_n)) \leq \ln (\frac{\sum_{p_i \in \mathcal{N}, p_i \neq p_n}f(p_i,q_n)}{|\mathcal{N}_{q_n}|})
$$
Given that:
\begin{equation*}
    f(p_i,q_n)=\frac{e^{sim(p_n^+,q_n)}}{\sum_{p_i^- \in \mathcal{N}_{q_n}}e^{sim(p_i^-,q_n)}+e^{sim(p_n^+,q_n)}}
\end{equation*}, we can further get:
\begin{equation*}
    \begin{aligned}
        &\frac{1}{|\mathcal{N}_{q_n}|}\sum_{p_i \in \mathcal{N}, p_i \neq p_n}\ln(f(p_i,q_n)) \leq \ln (\frac{1-f(p_n,q_n)}{|\mathcal{N}_{q_n}|})
    \end{aligned}
\end{equation*}
And so, we will have the:
\begin{equation}
    \begin{aligned}
        & f(p_n,q_n) \geq 1/(|\mathcal{N}_{q_n}|+1)\\
        \Leftrightarrow &|\mathcal{N}_{q_n}|*f(p_n,q_n) \geq 1-f(p_n,q_n)\\
        \Leftrightarrow &\ln(f(p_n,q_n)) \geq \ln (\frac{1-f(p_n,q_n)}{|\mathcal{N}_{q_n}|})\\
         \Rightarrow &\ln(f(p_n,q_n)) \geq \frac{1}{|\mathcal{N}_{q_n}|}\sum_{p_i \in \mathcal{N}, p_i \neq p_n }\ln(f(p_i,q_n))
    \end{aligned}\tag{2-2}
\end{equation}
Together with (2-1) and (2-2), we will get:
\begin{equation*}
    f(p_n,q_n) \geq 1/(|\mathcal{N}_{q_n}|+1) \Leftrightarrow (p_n,q_n) \textit{ will NOT be selected} 
\end{equation*}
Therefore, one sufficient condition of $(p_n,q_n)$ to be not selected is that the model give $p_n$ the higher score than random guess, i.e. $f(p_n,q_n) \geq 1/(|\mathcal{N}_{q_n}|+1)$
\end{proof}

\end{document}